%% file: main.tex
\documentclass[sigconf]{acmart}

\usepackage{CJKutf8}

\usepackage{amssymb}
\usepackage{amsthm}
\usepackage{algorithm}
\usepackage{enumitem}
\usepackage{picinpar}
\usepackage{algorithmicx}
\usepackage[noend]{algpseudocode}
\usepackage{subfigure}
\usepackage{multirow}
\usepackage{color}
\usepackage{balance}
\usepackage{enumitem}
\usepackage{hhline}
\usepackage[normalem]{ulem}

\usepackage{wrapfig}
\usepackage{bm}

\newcommand{\bL}{\ensuremath{\mathcal{L}}}

\newcommand{\bV}{\ensuremath{\mathcal{V}}}

\newcommand{\bN}{\ensuremath{\mathcal{N}}}

\renewcommand{\vec}[1]{\ensuremath{\mathbf{#1}}}
\newcommand{\stitle}[1]{\vspace{2mm} \noindent {\bf #1}}
\newcommand{\eg}{{\it e.g.}}
\newcommand{\etal}{{\it et al.}}
\newcommand{\ie}{{\it i.e.}}
\newcommand{\etc}{{\it etc.}}
\newcommand{\wrt}{w.r.t. }

\newcommand{\method}[1]{\textsf{#1}}
\newcommand{\model}{\method{HINormer}}

\newcommand{\eat}[1]{}

\AtBeginDocument{%
  \providecommand\BibTeX{{%
    \normalfont B\kern-0.5em{\scshape i\kern-0.25em b}\kern-0.8em\TeX}}}

\copyrightyear{2023}
\acmYear{2023}
\setcopyright{acmlicensed}\acmConference[WWW '23]{Proceedings of the ACM Web Conference 2023}{May 1--5, 2023}{Austin, TX, USA}
\acmBooktitle{Proceedings of the ACM Web Conference 2023 (WWW '23), May 1--5, 2023, Austin, TX, USA}
\acmPrice{15.00}
\acmDOI{10.1145/3543507.3583493}
\acmISBN{978-1-4503-9416-1/23/04}

\begin{document}

\title{HINormer: Representation Learning On Heterogeneous Information Networks with Graph Transformer}


\author{Qiheng Mao$^{1,4}$, Zemin Liu$^{2\dagger}$, Chenghao Liu$^{3\dagger}$, Jianling Sun$^{1,4}$}
 \affiliation{%
  \institution{$^{1}$Zhejiang University; $^{2}$National University of Singapore; \\
 $^{3}$Salesforce Research Asia; 
$^{4}$Alibaba-Zhejiang University Joint Institute of Frontier Technologies}
 \country{
 \{maoqiheng, sunjl\}@zju.edu.cn, zeminliu@nus.edu.sg, chenghao.liu@salesforce.com
 }
}
% \affiliation{
%  \{maoqiheng, sunjl\}@zju.edu.cn, zeminliu@nus.edu.sg, chenghao.liu@salesforce.com
%  }

\thanks{
    $^{\dagger}$The corresponding authors.
}

\renewcommand{\shortauthors}{Qiheng Mao, Zemin Liu, Chenghao Liu, Jianling Sun}

\begin{abstract}
%Heterogeneous graph neural networks (HGNNs) have been powerful deep learning architectures in heterogeneous graph representation learning, and heterogeneous Message-Passing mechanism within original heterogeneous graphs or meta-path based subgraphs is the key architecture of HGNNs consistent with homogeneous GNNs. However, more and more 
Recent studies have highlighted the limitations of message-passing based graph neural networks (GNNs), \eg, limited model expressiveness, over-smoothing, over-squashing, \etc\ To alleviate these issues, Graph Transformers (GTs) have been proposed which work in the paradigm that allows message passing to a larger coverage even across the whole graph. Hinging on the global range attention mechanism, GTs have shown a superpower for representation learning on homogeneous graphs. However, the investigation of GTs on heterogeneous information networks (HINs) is still under-exploited. In particular, on account of the existence of heterogeneity, HINs show distinct data characteristics and thus require different treatment. To bridge this gap, in this paper we investigate the representation learning on HINs with Graph Transformer, and propose a novel model named \model, which capitalizes on a larger-range aggregation mechanism for node representation learning. In particular, assisted by two major modules, \ie, a local structure encoder and a heterogeneous relation encoder, \model\ can capture both the structural and heterogeneous information of nodes on HINs for comprehensive node representations. We conduct extensive experiments on four HIN benchmark datasets, which demonstrate that our proposed model can outperform the state-of-the-art.
% Recent work has highlighted the limitation of Message-Passing based GNN models including limited model expressiveness, over-smoothing, over-squashing, \etc\ By allowing nodes to directly pass messages to all the others in the same graph, Graph Transformers (GTs) alleviate the above limitations. This global attention mechanism has shown great power in graph representation learning for homogeneous graphs, but it is still under-explored in the heterogeneous information networks (HINs). Different from homogeneous graphs, HINs mainly focus on node-level representation for large-scale networks, directly employing Graph Transformers is infeasible due to scalability issue. And how to integrate both structural and heterogeneous information, which are complex and important for HINs, into Graph Transformer architecture is full of challenge. To alleviate these concerns, in this paper we propose a novel heterogeneous Graph Transformer which achieves the goal of employing Graph Transformer on heterogeneous node representation learning with both structural and heterogeneous encoding. In particular, we proposes a light-weight and effective node-level Transformer architecture, attempt to enhance the local structural information with the neighborhood aggregation for Graph Transformer, and encode the heterogeneous semantics in a relative position encoding manner. We conduct extensive experiments on four HIN benchmark datasets, and demonstrate that our proposed model can outperform the state-of-the-art.
\end{abstract}

%    <concept>
%        <concept_id>10010147.10010257.10010293.10010294</concept_id>
%        <concept_desc>Computing methodologies~Neural networks</concept_desc>
%        <concept_significance>300</concept_significance>
%        </concept>
%    <concept>
%        <concept_id>10002950.10003624.10003633.10010917</concept_id>
%        <concept_desc>Mathematics of computing~Graph algorithms</concept_desc>
%        <concept_significance>500</concept_significance>
%        </concept>
%  </ccs2012>

% \ccsdesc[300]{Computing methodologies~Neural networks}
% \ccsdesc[500]{Mathematics of computing~Graph algorithms}





\maketitle

\input{sec-introduction}

\input{sec-related-work}

\input{sec-preliminaries}

\input{sec-model}

\input{sec-experiments}

\input{sec-conclusions}

\newpage
\balance
\bibliographystyle{ACM-Reference-Format}
\bibliography{main}

\appendix

\newpage
\balance
\input{sec-appendix}

\end{document}

%% file: sec-introduction.tex
\section{Introduction} \label{sec.intro}

Heterogeneous information networks (HINs) \cite{sun2011pathsim,sun2012mining,shi2016survey} are connecting structures with multiple types of nodes and edges, implying complicated heterogeneous semantic relations between nodes. 
As many real-world networks such as Web data can be modeled as HINs, they have attracted a surge of investigation which can further benefit the downstream graph analysis tasks. In particular, heterogeneous graph representation learning \cite{yang2020heterogeneous} arises as an effective tool to embed nodes into low-dimensional vectors by preserving both graph structures and heterogeneity, opening a great opportunity for HIN analysis.

Recently, more attention has been shifted to heterogeneous graph neural networks (HGNNs) \cite{wang2021self,zhao2022space4hgnn,lv2021we}, which generalize the key operation of neighborhood aggregation in graph neural networks (GNNs) \cite{wu2020comprehensive} onto HINs, with additionally modeling the graph heterogeneity. The capability of encoding both graph structure and heterogeneity gives rise to the outstanding performance of HGNNs in various HIN tasks, \eg, node classification and link prediction. Similar to GNNs, HGNNs also follow a message-passing paradigm to generate node representations, by recursively aggregating messages from the neighboring nodes in a heterogeneity-aware manner. 
In terms of heterogeneity utilization, HGNNs can be divided into two categories, \ie, meta-path based HGNNs and meta-path free HGNNs.
% The way of utilization of heterogeneity divides HGNNs into two categories,i.e., meta-path-based HGNNs and meta-path-free HGNNs. 
The former aggregates messages under the assistance of manually designed or automatically generated meta-paths \cite{sun2011pathsim, wang2019heterogeneous, LIU2022}, while the latter directly incorporates the heterogeneity information in representation learning without the help of meta-path \cite{hu2020heterogeneous, lv2021we,liu2017semantic}.
% The former aggregate messages under the assistance of handcraft meta-paths \cite{sun2011pathsim} firstly and then aggregate different heterogeneous semantic information; while the latter resorts to heterogeneity-aware parameters or modules to learn structural information and heterogeneous semantic information simultaneously, getting rid of meta-paths to extract heterogeneous information. \par

With the prevalent applications and in-depth investigation of GNNs, the limitations of message-passing based frameworks are beginning to emerge. 
Firstly, the expressive power of GNNs is bounded by the Weisfeiler-Lehamn isomorphism which means it may fail to learn finer-grained graph structures \cite{xu2018powerful}. 
Secondly, GNNs suffer from the over-smoothing problem, which limits the capacity of GNNs for building deeper architectures \cite{chen2020measuring}. 
Thirdly, the neighborhood-based aggregation impairs the ability of GNNs to propagate messages to distant nodes, and only short-range signals can be captured due to the over-squashing problem \cite{alon2020bottleneck}. 
% Thirdly, the graph fails to propagate messages flowing from distant nodes, and learns only short-range signals from the graph data due to over-squashing problem\cite{alon2020bottleneck}. 
Although there are some works that try to separately tackle these problems with GNN-backboned frameworks \cite{rong2019dropedge, zhao2019pairnorm, sankar2020beyond, sankar2022self}, the limitation of message-passing architectures for message passing still remains.
% under the architecture of GNNs \cite{rong2019dropedge, zhao2019pairnorm, sankar2020beyond, sankar2022self}, 
% there is still no complete solution that can effectively solve the above problems for message-passing paradigm. \par

The success of Transformer architectures in fields like natural language processing \cite{vaswani2017attention} and computer vision \cite{dosovitskiy2020image,liu2021swin, han2022survey}, has attracted a surge of attention from graph area, which accordingly demonstrates the strong capability of learning interactions between nodes with global attention mechanism.
% . Using graph structures as inputs, Transformer gradually began to demonstrate its strong modeling capability in graph fields by learning interactions of each node pair in a direct manner with global attention. 
A flurry of Graph Transformer models \cite{mialon2021graphit, wu2021representing} have been proposed to replace the message-passing paradigm and they have achieved the state-of-the-art performance on a variety of graph tasks \cite{ying2021Transformers, rampavsek2022recipe}.
% A flurry of Transformer-like Graph Transformer models (Graph Transformers)\cite{mialon2021graphit, wu2021representing} have been proposed to replace the message-passing paradigm and these models surpass the previous performance records achieved by GNNs on a variety of graph benchmarks\cite{ying2021transformers, rampavsek2022recipe}. 
Despite the success of Graph Transformers, there is still a significant limitation of applying Transformer architectures to graphs: global attention incurs quadratic complexity \wrt the number of nodes in a graph \cite{zhao2021gophormer,rampavsek2022recipe}, which results in the poor scalability toward large networks. Due to this issue, small-scale graph-level Transformers thrive while large-scale node-level Transformers do not. In particular, Transformer-based graph models for large-scale and heterogeneity-complex HINs are still under-exploited.

\stitle{Challenges and present work.}
To narrow this gap, in this paper we investigate the general application of graph Transformers on HINs for node representation learning, which is, however, a non-trivial problem.
On the one hand, though graph Transformers have strong advantages in the extraction of high-order features and long-range dependencies, the large HIN sizes usually give rise to the infeasibility of learning discriminative node representations due to their large perceptual fields, marginalizing the local-view structures of nodes \cite{wu2021representing,chen2022structure,rampavsek2022recipe, lv2021we}. So, \emph{how to design an efficient node-level Transformer encoder by virtue of the local-view structure information?} 
On the other hand, HINs typically preserve various type information which usually reflects the semantic relations between nodes. Therefore, a successful design of HIN Transformer should be capable of capturing the heterogeneity to express the diverse semantics. Thus, \emph{how to effectively capture the heterogeneous semantic relations between nodes with graph Transformer on HINs?}

To address these challenges, in this paper we propose a novel \underline{H}eterogeneous \underline{I}nformation \underline{N}etwork Transf\underline{ormer}, \ie, \model, 
% i.e., \underline{H}eterogeneous \underline{I}nformation \underline{N}etwork Transf\underline{ormer}, 
to investigate the feasibility of replacing HGNNs with more powerful Transformer architectures, free from pre-designed semantic schema like meta-paths, to learn both structure and heterogeneity simultaneously in an end-to-end manner. In particular, \model\ employs a lightweight and effective node-level self-attention mechanism, to capture the local structure information with context aggregation. Besides, it encodes the heterogeneous semantics on HINs with a novel relative relation encoder, by exploiting the semantic relations between nodes to benefit the expressiveness of HIN Transformer for heterogeneity preservation.
% which benefits the heterogeneous Graph Transformer from effective utilization of HIN information. 

Inspired by \cite{hamilton2017inductive, zhao2021gophormer}, to enable the node representation learning on HINs with the Transformer architecture, we propose a $D$-hop node context sampling approach to derive the contextual sequence for each node. In addition, we replace the dot-product global attention mechanism with a more lightweight GATv2 \cite{brody2021attentive} to save the learnable parameters in order to alleviate overfitting. 
% since we empirically observe that the lightweight attention-based GNNs can meet the performance requirement with fewer parameters (see Sect.~\ref{XXX}).
% and remove the feed-forward network from each Transformer layer to alleviate the risk of overfitting. 
Furthermore, Lv \etal\ \cite{lv2021we} empirically observe that the lightweight homogeneous GNNs can meet the performance requirement for representation learning on HINs with even fewer parameters.
Therefore, \emph{to address the first challenge}, drawing inspiration from this, as well as the successful practice on homogeneous graph Transformers, we propose a novel local structure encoder to extract the contextual structures on HINs, which is lightweight in structure and capable of recursively aggregating contextual information for a target node on the HIN. The enhanced node features are further employed as the input of heterogeneous Transformer to complement the necessary local-view structure information for discriminative node-level representation learning.
% The SimpleHGN \cite{lv2021we} points the performance of simple homogeneous GNNs like GCN, GAT are competitive even better than the existing HGNNs that depend on meta-paths to capture high-order semantic information or overemphasize the importance of heterogeneity. 
%Aggregating neighborhood information and generating local-view representations is not only important in homogeneous graphs\cite{wu2021representing,chen2022structure,rampavsek2022recipe}, but also in the complex HINs. 
% Inspired from the crucial findings of SimpleHGN and current successful practices made on homogeneous Graph Transformers, we propose a novel local structural information encoding to extract the local structures from original HINs. The local structural information encoding recursively aggregate neighborhood information on the original HIN with a heterogeneous feature projector. The enhanced node features are used as the input of heterogeneous Transformer to complement the necessary local-view structural information for discriminative node-level representations.

\emph{To address the second challenge}, given a target node, we assign each contextual node with a relational encoding to express the semantic relations with the target one, which can further provide heterogeneous information for Transformer in the contextual aggregation process.
To achieve this, we propose a novel heterogeneous relation encoder to learn the relational encodings in order to express the semantic proximities between nodes. The heterogeneous relation encoder takes both heterogeneous information and network structure as inputs, recognizes the importance of different types, and updates relational encodings with the learned type-aware weights from the neighborhood. In particular, the relation encoder concentrates on the relative proximities between nodes on the HIN without taking into account node features, in contrast to the local structure encoder which focuses on the extraction of local structures with the assistance of node features. The derived heterogeneous relational encodings can reflect the semantic proximities between nodes in the heterogeneity view, which is also necessary for Graph Transformer on HINs. \model\ will further employ this relative relational encoding in the global self-attention mechanism as the attention map bias \cite{rampavsek2022recipe} to enhance the awareness ability of heterogeneity.

\stitle{Contributions.} 
To the best of our knowledge, this is the first work to investigate the application of global-attention based Transformer on heterogeneous information networks.
% To the best of our knowledge, this is the first heterogeneous Graph Transformer to investigate the application of global-attention based Transformer architecture on HINs. 
In summary, our contributions are three-fold. 
(1) We design a new representation learning paradigm for HINs by innovatively applying Graph Transformer for node embeddings.
% architecture for node-level representation learning on HINs and propose an effective paradigm.
(2) We propose a novel model \model, which consists of a local structure encoder and a heterogeneous relation encoder to capture both structure and heterogeneity on HINs to assist the Transformer for node representation learning.
(3) Extensive experiments on four benchmark datasets demonstrate that \model\ can outperform the state-of-the-art baselines.

%% file: sec-related-work.tex
\section{Related work}

\stitle{Heterogeneous Network Embedding.}
The development of graph embedding approaches \cite{perozzi2014deepwalk,tang2015line,grover2016node2vec} opens great opportunities for graph analysis, which aim to map nodes or substructures into a low-dimensional space in which the connecting structures on the graph can be preserved. However, real-word networks are often composed of multi-modal and multi-typed nodes and relations \cite{yang2020heterogeneous}, which brings significant research attention on heterogeneous network embeddings (HNEs) \cite{shi2016survey, wang2022survey}. 
%The basic goal of network embedding is to capture network topological information, the challenge to heterogeneous network embedding is the complex interactions among multi-typed nodes with multi-typed links. 
Similar to graph embedding approaches, there are two major categories of HNEs, \ie, random walk approaches \cite{dong2017metapath2vec,fu2017hin2vec,liu2017semantic} and first/second-order proximity methods \cite{tang2015pte,shi2018easing}. Meta-paths or type-aware network closeness constraints are used to exploit network heterogeneity for HNEs. All of these methods are considered as shallow network embedding methods and can not preserve attribute information of nodes.

\stitle{Heterogeneous Graph Neural Network.}
%Compared with shallow graph embedding methods \cite{cai2018comprehensive}, Graph Neural Networks can preserve both the surrounding structure and content information simultaneously into node embeddings, which have attracted considerable attention in the field of graph representation learning \cite{wu2020comprehensive}. 
The success of GNNs on homogeneous, which preserve both the surrounding structure and content information simultaneously into node embeddings \cite{cai2018comprehensive, wu2020comprehensive}, has led to a flurry of proposed heterogeneous graph neural networks in recent years \cite{lv2021we,wang2021self}. Similar to HNEs, the flexibility of HGNNs model design mainly lies in the modeling of heterogeneous information (Heterogeneity). The way of utilization of heterogeneity divides HGNNs into two categories, \ie, meta-path based HGNNs \cite{wang2019heterogeneous, fu2020magnn} and meta-path free HGNNs \cite{zhu2019relation, hu2020heterogeneous}. Meta-path based HGNNs utilize meta-paths \cite{huang2016meta} to aggregate information from type-specific neighborhood, and is able to capture higher-order semantic information specified by the selected meta-paths. 
The selection of meta-paths needs expert knowledge of networks and plays a key role in model performance. Meta-path free HGNNs get rid of dependence on handcraft meta-paths, they employ message-passing mechanism directly on the original heterogeneous network with node/edge type-aware module, so that the model can capture structural and semantic information simultaneously. The most relevant HGNN to us is \method{HGT} \cite{hu2020heterogeneous}. Extended from \method{GAT}, \method{HGT} calculates heterogeneous attention scores for 1-hop neighboring nodes w.r.t. edge types, which still follows message-passing paradigm and suffers from the limitation of short-range view.
%To some extent, the success of meta-path free HGNNs proves that meta-path is not the only way for HGNNs.\par
%The two HGNNs that are most closely related to us are GTN\cite{yun2019graph} and HGT\cite{hu2020heterogeneous}. GTN doesn't relate to Transformer architecture, but only concerns how to automatically learn valuable meta-paths on HINs. 

\stitle{Graph Transformers.}
Recently, a growing body of research has highlighted the limitations of Message-Passing-based GNN models such as expressiveness bounds, over-smoothing, over-squashing and so on \cite{xu2018powerful,chen2020measuring,alon2020bottleneck}. 
Increasing attention from graph representation learning field has been paid to the design of Graph Transformers \cite{dwivedi2020generalization,kreuzer2021rethinking,wu2021representing,ying2021transformer}. The fully-connected global attention mechanism has shown its power in many different graph representation learning tasks. The generalizations of \method{Transformer} to graph-structured data are mainly around integration of graph structural information into Transformer architectures. Some Graph Transformers resort to combination with GNNs to capture local structure information, like \method{GraphTrans} \cite{wu2021representing}, \method{GraphiT} \cite{mialon2021graphit}. Some propose to use graph positional encoding and structural encoding to complement topological information into Graph Transformer \cite{dwivedi2020generalization, kreuzer2021rethinking} like positional encoding used in the original \method{Transformer} \cite{vaswani2017attention} and its variants in different fields \cite{dosovitskiy2020image}. 
%For example, GT\cite{dwivedi2020generalization}  utilize Laplacian eigenvectors to represent positional encodings and fuse them with the original features of nodes as the input of Graph Transformer, while SAN\cite{kreuzer2021rethinking} leverage the full spectrum of Laplacian matrix to learn the positional encodings. 
And other graph structural information like the shortest path length, node degree and edge features \cite{ying2021transformer} can also be used as position encodings to enhance the expressiveness of Graph Transformer in structural aspect. But most existing Graph Transformers directly use the whole graph as input, making them hard to handle large-scale real-world network like HINs because the global attention mechanism has square complexity w.r.t. number of nodes \cite{zhao2021gophormer}.

%% file: sec-preliminaries.tex
\section{Preliminaries}

\subsection{Heterogeneous Information Network}

A heterogeneous information network (HIN) can be defined as $G = \{V,E, \vec{X}, \phi,\psi\}$,
where $V$ is the set of nodes, $E$ is the set of edges, and $\vec{X}\in\mathbb{R}^{|V|\times d_x}$ is the feature matrix, with $\vec{x}_v\in\mathbb{R}^{d_x}$ denoting the feature vector of node $v\in V$. Each node $v\in V$ has a type $\phi(v)$ and each edge $e\in E$ has a type $\psi(e)$, where $\phi$ is node type mapping function and $\psi$ is edge type mapping function, respectively. The sets of
node types and edge types are denoted as $T_v = \{\phi(v):\forall v \in V\}$
and $T_e = \{\psi(e):\forall e \in E\}$, respectively. When $|T_v|=|T_e|=1$, the graph degenerates into a homogeneous graph.

\subsection{Graph Neural Networks}
GNNs \cite{wu2020comprehensive,liu2021tail,liu2021nodewise} and HGNNs \cite{wu2020comprehensive, yang2020heterogeneous} typically depend on the key operation of layer-wise neighborhood aggregation to recursively pass and transform messages from the neighboring nodes to form the representation of the target node. Formally, let $\phi_g(\cdot;\theta_g)$ denote a GNN architecture parameterized by $\theta_g$. In the $l$-th layer, the representation of node $v$, \ie, $\vec{h}^l_v\in\mathbb{R}^{d_l}$, can be calculated by
\begin{equation}\label{eq.gnn}
    \vec{h}_v^l = \textsc{Aggr}(\vec{h}_v^{l-1},\{\vec{h}_i^{l-1}: i \in \bN_v\}; \theta^l_g).
\end{equation}
where $\bN_v$ is the neighbors set of node $v$ (or type-specific neighborhood for HGNNs), and $\textsc{Aggr}(\cdot;\theta^l_g)$ is the neighborhood aggregation function parameterized by $\theta^l_g$ in layer $l$. Given a total of $\ell$ layers, $\theta_g=\{\theta^1_g, \theta^2_g, \ldots, \theta^{\ell}_g\}$ denotes all the parameters set in the aggregation model. 
In addition, different GNN architectures may differ in the neighborhood aggregation function, \eg, mean-pooling aggregation in \method{GCN} \cite{kipf2016semi}, attention-based aggregation in \method{GAT} \cite{velivckovic2017graph}. 
In particular, our local structure encoder is agnostic to the base GNN models, and we aim to make it flexible to most neighborhood aggregation based GNNs.

\subsection{Transformer Architecture} \label{sec.original-transformer}

The standard Transformer layer is composed of two major components, a multi-head self-attention (MSA) module and a feed-forward network (FFN). In the following part, we will briefly introduce MSA without multi-head for simplicity.

Given an input sequence $\vec{H} = [\vec{h_1},\vec{h}_2,...,\vec{h}_n]^\top \in \mathbb{R}^{n \times d} $, where $d$ is the hidden dimension and $\vec{h}_i \in \mathbb{R}^{d} $ is the hidden representation at position $i$, the MSA firstly projects the input $\vec{H} $ to the query-, key-, value-spaces, denoted as $\vec{Q}, \vec{K}, \vec{V} $, by resorting to three parameter matrices $\vec{W_{Q}} \in \mathbb{R}^{d \times d_{K}}$, $\vec{W_{K}} \in \mathbb{R}^{d \times d_{K}} $ and $\vec{W_{V}} \in \mathbb{R}^{d \times d_{V}} $, as
\begin{equation}\label{eq.qkv}
    \vec{Q} = \vec{HW_{Q}}, \quad \vec{K} = \vec{HW_{K}}, \quad \vec{V} = \vec{HW_{V}}.
\end{equation}
Then, the scaled dot-product attention mechanism is applied to the corresponding $<\vec{Q},\vec{K},\vec{V}>$ as
%Then, in each head $\mathtt{h} \in \{1,2,..,n_h\}$, the scaled %dot-product attention mechanism is applied to the corresponding $<\bm{Q_{\mathtt{h}}},\bm{K_{\mathtt{h}}},\bm{V_{\mathtt{h}}}>$: 
\begin{equation}\label{eq.head_attention}
    \text{MSA}(\vec{H}) = \textsc{Softmax} \frac{\vec{Q}\vec{K}^{\top}}{\sqrt{d_K}}\vec{V}.
\end{equation}
%and the outputs from different heads are further concatenated %and transformed to obtain the final output of MSA:
%\begin{equation}\label{eq.MSA_output}
%    MSA(\bm{H}) = Concat(head_1,...,head_{n_h})\bm{W_O}
%\end{equation}
%where $\bm{W_O} \in \mathbb{R}^{d \times d}$ is a parameter matrix. In practice, we set $d_K = d_V = d / n_h$.
Thereafter, the output of MSA will be connected to the FFN with two layers of Layer Normalization (LN) \cite{ba2016layer} and the residual connection \cite{he2016deep} to obtain the output of the $l$-th Transformer layer, denoted as $\vec{H}^l$:
\begin{equation}\label{eq.final_output}
    \begin{aligned}
    & \widetilde{\vec{H}}^l = \text{LN}(\text{MSA}(\vec{H}^{l-1})+\vec{H}^{l-1}),   \\
    & \vec{H}^l = \text{LN}(\text{FFN}(\widetilde{\vec{H}}^l) + \widetilde{\vec{H}}^l).
    \end{aligned}
\end{equation}

By stacking with $L$ layers, Transformers could learn the feature-based proximities between different positions for the input sequences, then the final output $\vec{H}^L \in \mathbb{R} ^ {n \times d}$ can be used as the representation of the input sequence for downstream tasks.

\begin{figure*}[t]
\centering
\includegraphics[scale=0.6]{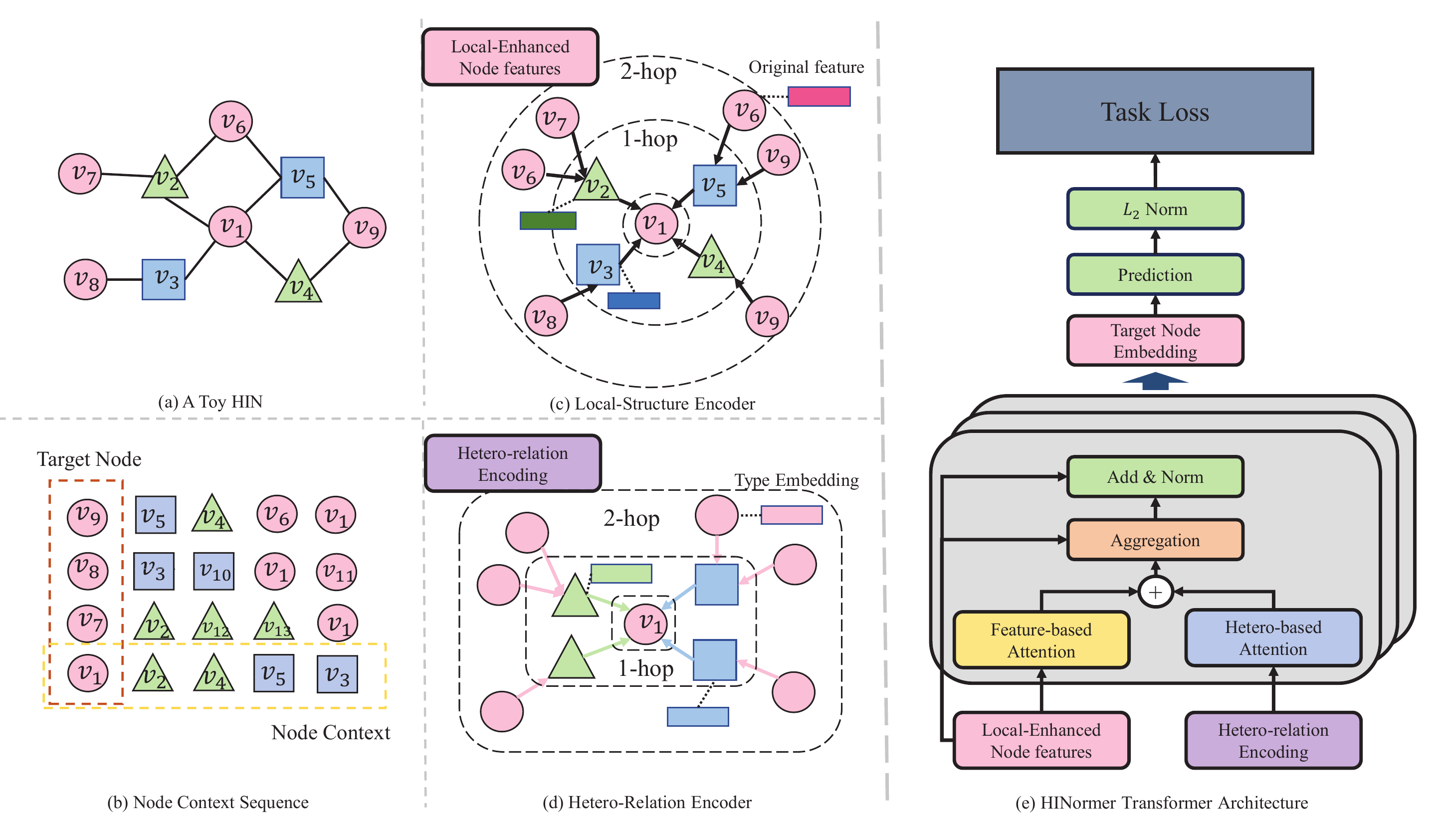}
\vspace{-3mm}
\caption{Overall framework of \model.}

\vspace{-3mm}
\label{fig.framework}
\end{figure*}

%% file: sec-model.tex
\section{The Proposed Model: \model}

% \subsection{Overall framework}
In this section, we present the concrete description of \model\ to explore the feasibility of applying more powerful Transformer architectures to node-level representation learning on HINs. We illustrate the overall framework of \model\ in Fig.~\ref{fig.framework}. Given a heterogeneous graph $G$ with full adjacency matrix $A$, we first adopt local structure encoder in Fig.~\ref{fig.framework}(c) and heterogeneous relation encoder in Fig.~\ref{fig.framework}(d) to capture the feature-based local structural information and heterogeneity-based semantic proximities for each node, respectively. The outputs of these two encoders are employed as the node features and relative positional encoding of the main part of the heterogeneous Graph Transformer, respectively. After the $D$-hop context sampling process for each node in Fig.~\ref{fig.framework}(b), \model\ is able to capture high-order semantic relations for the specific context with global self-attention mechanism and generate discriminative node-level representations by the integration of features, structures, and heterogeneity (Fig.~\ref{fig.framework}(e)). Finally, the node labels are predicted by the normalized node representations generated by Transformer layers in Fig.~\ref{fig.framework}(e) and the model is trained with a supervised classification loss. In the following part, we will illustrate the details for each part.

\subsection{Node-level Heterogeneous Graph Transformer Architecture} \label{sec.objective}

%对于Node-level的描述再简单明了一些
Current Graph Transformers directly utilize the entire graph to generate input sequences \cite{kreuzer2021rethinking,mialon2021graphit, wu2021representing}, making these approaches only applicable to small datasets (\eg, small graphs) due to expensive memory cost. In our setting of node representation learning on HINs, the given HIN usually has a large size (\eg, the number of nodes), so in general it is impossible to treat all nodes as one input sequence. In addition, the typical tasks on HINs are node-level, \eg, node classification and link prediction, where only one graph is in general involved. Consequently, existing Graph Transformer methods are not applicable for this kind of data settings and fail to learn expressive node-level representations \cite{zhao2021gophormer,rampavsek2022recipe}.

In order to realize the node representation learning on HINs with the Transformer architecture, inspired by \cite{hamilton2017inductive, zhao2021gophormer}, we propose the $D$-hop node context sampling process to get the input sequences for each node. More precisely, given a target node, we add the neighbors of each hop to iteratively construct the input sequence with pre-defined maximum depth $D$. Besides, in order to unify the sequence length of different nodes, all input sequences will be trimmed to a fixed length $S$ which is compatible with global attention mechanism.  Employing global attention mechanism in the local context emphasizes more on local structural information which is crucial to the discrimination of node-level representations and alleviates the deterioration of noise from the too-long-range nodes. In addition, we also propose a local structure encoder to capture structural information from the context of each node, which we will detailedly illustrate in Sect.~\ref{sec.local-struct}.

Formally, the sampled sequence of a target node $v$ is denoted as $s(v) = \{v,v_1,...,v_{S-1}\}$, and the corresponding input for Transformer is denoted as $\vec{H}_v^s = [\vec{h}_v^s, \vec{h}_{v_1}^s, ..., \vec{h}_{v_{S-1}}^s]^\top \in \mathbb{R} ^ {S \times d}$, where $d$ is the node embedding dimension and $\vec{h}_{v_i}^s$ is the local-enhanced node features from local structure encoder. In total, the entire input of Graph Transformer is denoted by $\vec{H}^s = [\vec{H}_1^s,\vec{H}_2^s,...\vec{H}_N^s]^\top \in \mathbb{R} ^ {N \times S \times d}$.

After the encoding of the $L$-layer Transformer, we achieve the output $\vec{H}^L = [\vec{H}_1^L, \vec{H}_2^L, ..., \vec{H}_N^L]^\top \in \mathbb{R} ^ {N \times S \times d}$, then we require an additional step to get the representation of the target node, which can be achieved by a readout function as
\begin{equation}\label{eq.readout}
    \vec{h}_v = \textsc{Readout}(\vec{h}_{s(v)[i]}^{L}|v_i \in s(v)),
\end{equation}
where $s(v)[i]$ denotes the $i$-th node in $s(v)$.
In practice, we directly use the target node representation in the sequences as the output node embeddings, \ie, $\vec{h}_v = \vec{h}_{s(v)[0]}^{L}$, which we observe can exceed the performance of other graph readout functions \cite{ying2018hierarchical,xu2018powerful}.

The standard Transformer architecture usually contains much more learnable parameters than HGNNs, which raises difficulties for effective model training with limited supervision, and might give rise to the overfitting issue.
% makes it hard to get sufficient training under a semi-supervised setting with scarce node labels and easy to overfitting. 
To further improve the adaptability of Transformer to node-level representation learning on HINs, we propose two simple yet effective changes to Transformer architecture to prevent the risk of overfitting. 
%GATv2 公式补充，简单分析
First, instead of the original self-attention mechanism which involves more learnable parameters as illustrated in Sect.~\ref{sec.original-transformer}, we propose to utilize \method{GATv2} \cite{brody2021attentive} as the replacement.
% to replace dot-product self-attention as global attention mechanism. 
\method{GATv2} is proven to be a universal approximator attention function which is more powerful than dot-product self-attention in Transformer yet with fewer parameters \cite{brody2021attentive}. We provide the differences between \method{GAT} \cite{velivckovic2017graph} and \method{GATv2} \cite{brody2021attentive} as follows.
\begin{equation}\label{eq.gat}
    \text{\method{GAT}}:    \alpha(\vec{h}_i,\vec{h}_j) = \textsc{LeakyReLU}(\vec{a}^{\top}\cdot[\vec{W}\vec{h}_i||\vec{W}\vec{h}_j]),
\end{equation}
\begin{equation}\label{eq.gatv2}
    \text{\method{GATv2}}:  \alpha(\vec{h}_i,\vec{h}_j) = \vec{a}^\top\cdot\textsc{LeakyReLU}(\vec{W}\cdot[\vec{h}_i||\vec{h}_j]).
\end{equation}
Second, we delete the feed-forward network (FFN) in each Transformer layer, which can greatly reduce the number of learnable parameters and have no significant negative impact on model performance. Note that, here the inexistence of FFN in heterogeneous Graph Transformer further demonstrates the effectiveness of the global attention mechanism. 

Overall, the calculation of a Transformer layer in \model\ is formulated as
\begin{equation}\label{eq.transformer_layer}
    \vec{H}^{l} = \text{LN}(\vec{H}^{l-1} + \textsc{HeteroMSA}(\vec{H}^{l-1},\vec{R})),
\end{equation}
where $\textsc{HeteroMSA}(\cdot)$ is the heterogeneity-enhanced multi-head self-attention and $\vec{R}$ represents the heterogeneous relational encodings, which we will introduce in detail in Sect.~\ref{sec.RE}.

\subsection{Local Structure Encoder} \label{sec.local-struct}

 Existing representative studies of HGNNs, \eg, \method{SimpleHGN} \cite{lv2021we}, have proved the importance of structural information from a local view for heterogeneous graph representation learning, and the homogeneous graph neural networks such as \method{GCN} \cite{kipf2016semi} and \method{GAT} \cite{velivckovic2017graph}, can achieve competitive performance even without the modeling of heterogeneity on HINs. Aggregating features from the neighborhood is a direct way to involve local structural information on HINs \cite{chen2022structure,rampavsek2022recipe}, and the heterogeneity can also be learned by the heterogeneous feature projection \cite{wang2021self,lv2021we}. Then the achieved local structure encoding will serve as the input of the heterogeneous Graph Transformer as a local sub-structure enhanced feature to benefit the global attention mechanism.

\stitle{Heterogeneous feature projection.}
As the features of different types of nodes on HINs usually exist in different feature spaces, the first priority is to map their features into a shared feature space. We employ a linear transformation with bias term as the heterogeneous feature projection function, as
\begin{equation}\label{eq.feat_pro}
    \vec{h}_v = \vec{W}_{\phi(v)}\vec{x}_v + \vec{b}_{\phi(v)},
\end{equation}
where $\vec{W}_{\phi(v)} \in \mathbb{R}^{d \times d_x}$ is a learnable parameter matrix, and $\vec{b}_{\phi(v)} \in \mathbb{R}^{d}$ is a learnable bias term.
% , $d_{\phi(v)}$ is original feature dimension of type $\phi(v)$ and $\vec{x}_v$ is original feature of node $v$.
Then the features of different types can be aggregated with the same function, and the heterogeneity is also preserved in the shared feature space.

\stitle{Local structure encoder.}
Since Transformer can only capture the feature-based similarity between nodes, the structural similarity and more important local context information for node-level representation learning are out of reach for the global attention mechanism. With an effective local structure encoder, global attention is able to consider the similarity between node pairs from more comprehensive views and thus can generate more representative embeddings.

% For the sampled sequence $s(v)$ of target node $v$, 
For a target node $v$, aggregation of neighborhood information with $K_s$ times can collect both structures and features of the contextual $K_s$-hop ego-graph of the target node, and the output represents the $K_s$ -subtree structure rooted at target node $v$. With an appropriate $K_s$ (usually very small), the over-smoothing and over-squashing problem can be alleviated. We implement the local structure encoder in the form of GNN-based neighborhood aggregation due to its effectiveness and computational efficiency, as follows.
\begin{equation}\label{eq.local_gnn}
    \vec{h}_v^l = \textsc{Aggr}(\vec{h}_v^{l-1},\{\vec{h}_i^{l-1}: i \in \bN_v\}; \theta^l_s).
\end{equation}
Note that, this local structure encoder is agnostic to most GNN or HGNN architectures \cite{you2020design, zhao2022space4hgnn}, and a simple GNN such as \method{GCN} \cite{kipf2016semi} and \method{GraphSAGE} \cite{hamilton2017inductive} can bring significant performance improvement. In practice, we also implement the local structure encoder in a non-parametric way for the consideration of higher efficiency, as follows.
\begin{equation}\label{eq.local_adj}
%    \vec{h}_v^{K_s} = [\hat{\bm{A}}^{K_s}\bm{H}]_v,
\vec{h}_v^{K_s} = (\hat{\vec{A}}^{K_s}\vec{H})[v,:],
\end{equation}
where $\hat{\vec{A}} = \vec{D}^{-1/2}\vec{A}\vec{D}^{-1/2}$ is the normalized adjacency matrix which is used in \method{GCN} \cite{kipf2016semi}, and $\vec{Z}[i,:]$ denotes the $i$-th row in matrix $\vec{Z}$.
Then we directly use the output of the $K_s$-th layer, i.e, $\vec{H}^{s} = \vec{H}^{K_s}$, as the input of heterogeneous Graph Transformer. Please refer to Table~\ref{table.node-classification-gnns} in Experiments for the empirical evaluation of different GNN-backboned local structure encoders.

\subsection{Heterogeneous Relation Encoder}\label{sec.RE}

HINs have abundant heterogeneity to reflect node semantic proximities in a heterogeneous view.
% which is the specific structural relation between nodes on HINs. 
To tackle the lack of heterogeneity modeling for heterogeneous Graph Transformer, we propose a novel heterogeneous relation encoder to learn the heterogeneous semantic proximities between nodes.

To encode the heterogeneity of HINs, for a target node $v$, we use the one-hot type vector as the initialization of heterogeneous relational encoding $\vec{r}_{v}^{0} = \vec{T}[\phi(v),:]$, where $\vec{T} \in \mathbb{R}^{|T_v| \times |T_v|}$ is the one-hot type embedding.
% and $\phi()$ is node type mapping function. 
As $|T_v|$ is usually a small value, the relation encoder is lightweight and fast to calculate.

In particular, we implement heterogeneous relation encoder in the way of neighborhood aggregation to encode the local heterogeneity and structures, and the number of layers is denoted as $K_h$.
The aggregation function of relation encoder is model-agnostic, we can use a simple GNN aggregation function. In practice we employ \method{GCN} as the aggregation function due to its simplicity and effectiveness. And we introduce additional learnable type-aware weights to adjust the influence of different types in each aggregation step since different types may contribute differently.
% usually take imbalanced effect to the representations of the target node. 
This simple heterogeneous model further improves the ability of \model\ to capture heterogeneity, and the calculation of relational encoding of node $v$ in iteration step $t$, \ie, $\vec{r}_{v}^{t}$ can be formalized as 
\begin{equation}\label{eq.re_agg}
    \vec{r}_{v}^{t} = \sum_{u \in N(v)}w_{\phi(u)}^{t-1}f(\vec{r}_{u}^{t-1};\theta^t_h),
\end{equation}
where $w_{\phi(u)}$ is the learnable aggregation weight of type $\phi(u)$, and $f(\cdot;\theta^t_h)$ is the transformation function parameterized by $\theta^t_h$. 
With $K_h$ iterations, the output $\vec{r}^{K_h}_{v}$ represents the contextual heterogeneity of node $v$, by comparing which we can derive the heterogeneity-aware proximities between two nodes.
% With type-aware aggregation of $K_h$ step, generated heterogeneous relation encoding $\vec{r}_{v}$, \ie, $\vec{r}^{K_h}_{v}$ will capture comprehensive heterogeneous semantic relation information of node $v$ on HINs. 
Then we can employ it as the heterogeneity-aware positional encoding of node $v$ and feed into $\textsc{HeteroMSA}(\cdot)$, to adjust the self-attention mechanism to capture the semantic proximities between nodes. 
% In this way, our \model\ can encode more heterogeneous information for learning more sophisticated global relation between context nodes. 
Thus, the attention in $\textsc{HeteroMSA}(\cdot)$ can be further modified as follows,
\begin{equation}\label{eq.attmap_re}
    \begin{aligned}
    & \vec{q}_i^R = \vec{W}_{\vec{Q}_R}\vec{r}_{i}, \quad \vec{k}_j^R = \vec{W}_{\vec{K}_R}\vec{r}_{j},   \\
    & \hat{\alpha}_{i,j} = \alpha_{i,j} + \beta \cdot \vec{q}_i^R \vec{k}_j^R,
    \end{aligned}
\end{equation}
where $\alpha_{i,j}$ is the attention score calculated by feature-based attention mechanism (corresponding to the $\vec{H}^{l-1}$ in Eq.~\eqref{eq.transformer_layer}),  $\vec{W}_{\vec{Q}_R},\vec{W}_{\vec{K}_R} \in \mathbb{R}^{|T_v| \times |T_v|}$ are the learnable projection matrices, and $\beta$ is the hyperparameter to modulate the weight of heterogeneous proximities. 
The heterogeneous attention is also extended to multi-head and we leave out here for simplicity. 
Finally, $\hat{\alpha}$ is further utilized in the attentive aggregation, \ie, $\textsc{HeteroMSA}(\cdot)$.
% \textbf{$\hat{\alpha}$ reflects comprehensive correlation between contextual nodes and is used in weighted aggregation of Graph Transformer.} 
% And we further add a temperature factor $\tau$ to adjust the sensitivity to a high value for heterogeneity-enhanced self-attention.

\subsection{Training objective}

Given the node-level representations on HINs, we feed them into the task loss for end-to-end training.
% In this section, we present the training objective of \model\ for node-level representation learning on HINs. 
We concentrate on node classification and leave other downstream tasks as future work. 
Formally, give the node representation $\vec{h}_v$ for node $v$, we further employ a linear layer $\phi_{\text{Linear}}(\cdot;\theta_{\text{pre}})$ parameterized by $\theta_{\text{pre}}$ to predict the class distribution as follows, 
% Given the target node representation $h_{v}$ learned by the \model, a linear prediction layer is adopted to predict the node class of $v$:
\begin{equation} \label{eq.pred_class}
    \tilde{\vec{y}}_{v} = \phi_{\text{Linear}}(\vec{h}_v;\theta_{\text{pre}}).
\end{equation}
where $\tilde{\vec{y}}_{v} \in \mathbb{R}^{C}$ is the prediction and $C$ is the number of classes. 
In addition, we further add an $L_2$ normalization on $\tilde{\vec{y}}_{v}$ for stable optimization \cite{ranjan2017l2,lv2021we}.
% In addition, we add an $L_2$ normalization \cite{ranjan2017l2,lv2021we} on the output of prediction network for stable optimization:
% \begin{equation} \label{eq.l2}
%     \tilde{\vec{y}_{v}} = \frac{ \tilde{\vec{y}_{v}}}{ ||\tilde{\vec{y}_{v}}||}.
% \end{equation}

Given the training nodes $\bV_{\text{tr}}$, for multi-class node classification, we employ cross-entropy as the overall loss, as
\begin{equation}    \textstyle \bL=\sum_{v\in\bV_{\text{tr}}}\textsc{CrossEnt}(\tilde{\vec{y}}_{v},\vec{y}_v),
\end{equation}
where $\textsc{CrossEnt}(\cdot)$ is the cross-entropy loss, and $\vec{y}_v\in\mathbb{R}^C$ is the one-hot vector that encodes the label of node $v$.
% \zemin{For single-label classification,  we use softmax and cross-entropy loss;  for multi-label datasets, we use a sigmoid activation and binary cross-entropy loss.}
Note that, for multi-label node classification, we can employ binary cross-entropy to calculate the overall loss.

%For single-label classification, the supervised node classification loss is achieved by the cross entropy loss of nodes in the traning data $V_{train}$:
%\begin{equation} \label{eq.cross_ent}
%     \bL_{single} = -\frac{1}{|V_{train}|}\sum_{v \in V_{train}}\textsc{CrossEnt}(\tilde{y_{v}},y_{v})
%\end{equation}
%where $y_{v} \in \mathbb{R}^{C \times 1}$ is the true label of node $v$. \\
%For multi-label classification, the supervised node classification loss is achieved by the binary cross-entropy loss with a sigmoid activation of $\tilde{y_{v}}$:
%\begin{equation} \label{eq.bin_cross_ent}
%     \bL_{multi} = -\frac{1}{|V_{train}|\times C}\sum_{v \in V_{train}}\sum_{i=1}^{C}\textsc{BinCrossEnt}(\sigma(\tilde{y_{v}})^{i},y_{v}^{i})
%\end{equation}
%We further present the algorithm for model training in Appendix~\ref{app.alg}.

%% file: sec-experiments.tex
\section{Experiments}
In this section, we conduct extensive experiments on node classification to evaluate the performance of the proposed \model, and further give detailed model analysis from several aspects.

\subsection{Experimental Setups}

\stitle{Datasets.} 
We employ four public HIN benchmark datasets, including two academic citation datasets (\emph{DBLP}\footnote{http://web.cs.ucla.edu/~yzsun/data/} \cite{lv2021we} and \emph{AMiner}\footnote{https://www.aminer.org/data/} \cite{Tang:08KDD}), 
a movie rating dataset (\emph{IMDB}\footnote{https://www.kaggle.com/karrrimba/movie-metadatacsv}), and a knowledge graph dataset (\emph{Freebase} \cite{bollacker2008freebase}). 
% and two movie-related datasets (\emph{IMDB}\footnote{https://www.kaggle.com/karrrimba/movie-metadatacsv} and \emph{Freebase} \cite{bollacker2008freebase}). 
% The academic citation datasets include \emph{DBLP}\cite{lv2021we}\footnote{http://web.cs.ucla.edu/~yzsun/data/}and \emph{AMiner}\cite{Tang:08KDD}\footnote{https://www.aminer.org/data/}, while the movie-related datasets include \emph{IMDB}\footnote{https://www.kaggle.com/karrrimba/movie-metadatacsv} and  \emph{Freebase}\cite{bollacker2008freebase}. 
Table~\ref{table.datasets} summarizes the statistic of the datasets. 
% \zemin{For node types without attributes, we use one-hot vectors as their original features.} 
For datasets without node features, we use the one-hot vectors with length $|V|$ as the node initial features to denote their existence.
We provide further details for the datasets in Appendix~\ref{app.datasets}.

\begin{table}[tbp]
\center
\small
\addtolength{\tabcolsep}{-0.6mm}
\caption{Summary of datasets.\label{table.datasets}}
\vspace{-2mm}
\resizebox{1.0\linewidth}{!}{
\begin{tabular}{@{}c|rrrrrr@{}}
\toprule
	 & \# Nodes & \# Node Types & \# Edges & \# Edges Types & Target & \# Classes    \\ 
\midrule
DBLP              & 26,128      & 4       & 239,566      & 6         & author            & 4   \\
IMDB               & 21,420     & 4    & 86,642     & 6         & movie            & 5  \\
Freebase      & 43,854   & 4      & 15,1034      & 6         & movie            & 3   \\
AMiner         & 55,783      & 3     &  153,676     & 4         & paper            & 4   \\
 \bottomrule
\end{tabular}}
\vspace{-2mm}
\end{table}

\begin{table*}[!t] 
    \centering
    \small
    \addtolength{\tabcolsep}{1.5pt}
    \caption{Performance evaluation on node classification, with GCN as local structure encoder.
    }
    \label{table.node-classification-all}%
    \vspace{-3mm} 
    {\footnotesize Henceforth, tabular results are in percent; the best result is \textbf{bolded} and the runner-up is \underline{underlined}. A dash (-) denotes that the models run out of memory on large graphs.
    }
    \\[2mm] 
    \begin{tabular}{@{}l|cc|cc|cc|cc@{}}
    \toprule
   \multirow{2}*{Methods} & \multicolumn{2}{c|}{DBLP} &
   \multicolumn{2}{c|}{IMDB} &  \multicolumn{2}{c|}{Freebase} &  \multicolumn{2}{c}{AMiner}\\ 
      & Micro-F1 & Macro-F1  & Micro-F1 & Macro-F1  & Micro-F1 & Macro-F1  & Micro-F1 & Macro-F1   \\\midrule\midrule
     \method{GCN}  &91.47 \textpm 0.34 &90.84 \textpm 0.32 &64.82 \textpm 0.64 &57.88 \textpm 1.18 & 68.34 \textpm 1.58 & 59.81 \textpm 3.04 & 85.75 \textpm 0.41 & \uline{75.74} \textpm 1.10\\
\method{GAT}    &93.39 \textpm 0.30 &93.83 \textpm 0.27 & 64.86 \textpm 0.43 &58.94 \textpm 1.35 & \uline{69.04} \textpm 0.58 & 59.28 \textpm 2.56 & 84.92 \textpm 0.68 & 74.32 \textpm 0.95\\ 
\method{Transformer} &93.99 \textpm 0.11 &93.48 \textpm 0.12 & 66.29 \textpm 0.69 &62.79 \textpm 0.65 & 67.89 \textpm 0.39 & \uline{63.35} \textpm 0.46 & 85.72 \textpm 0.43 & 74.15 \textpm 0.28\\
\midrule
\method{RGCN} &92.07 \textpm 0.50 &91.52 \textpm 0.50 &62.95 \textpm 0.15 & 58.85 \textpm 0.26
& 60.82 \textpm 1.23 &59.08 \textpm 1.44 & 81.58 \textpm 1.44 & 62.53 \textpm 2.31 \\
\method{HetGNN} &92.33 \textpm 0.41 & 91.76 \textpm 0.43 &51.16 \textpm 0.65 & 48.25 \textpm 0.67 & 62.99 \textpm 2.31 & 58.44 \textpm 1.99 & 72.34 \textpm 1.42 & 55.42 \textpm 1.45 \\
\method{HAN} &92.05 \textpm 0.62 &91.67 \textpm 0.49 & 64.63 \textpm 0.58 & 57.74 \textpm 0.96 & 61.42 \textpm 3.56 & 57.05 \textpm 2.06 & 81.90 \textpm 1.51 & 64.67 \textpm 2.21\\
\method{GTN} &93.97 \textpm 0.54 &93.52 \textpm 0.55 &65.14 \textpm 0.45 & 60.47 \textpm 0.98 & - & - & - & -\\
\method{MAGNN} &93.76 \textpm 0.45 &93.28 \textpm 0.51 &64.67 \textpm 1.67 &56.49 \textpm 3.20 & 64.43 \textpm 0.73 & 58.18 \textpm 3.87 &82.64 \textpm 1.59 & 68.60 \textpm 2.04 \\
\midrule
\method{RSHN} &93.81 \textpm 0.55 &93.34 \textpm 0.58 &64.22 \textpm 1.03 & 59.85 \textpm 3.21  & 61.43\textpm5.37 & 57.37 \textpm 1.49 & 73.33 \textpm 2.71 & 51.48 \textpm 4.20  \\
\method{HetSANN} &80.56 \textpm 1.50 &78.55 \textpm 2.42 & 57.68 \textpm 0.44 & 49.47 \textpm 1.21 & - & - & - & -\\
\method{HGT} &93.49 \textpm 0.25 &93.01 \textpm 0.23 & 67.20 \textpm 0.57 & 63.00 \textpm 1.19 & 66.43 \textpm 1.88 & 60.03 \textpm 2.21 & 85.74 \textpm 1.24 & 74.98 \textpm 1.61 \\
\method{SimpleHGN} &\uline{94.46} \textpm 0.22 &\uline{94.01} \textpm 0.24 &\uline{67.36} \textpm 0.57 & \uline{63.53} \textpm 1.36& 67.49 \textpm 0.97 & 62.49 \textpm 1.69 & \uline{86.44} \textpm 0.48 & 75.73 \textpm 0.97  \\
\midrule
\model &\textbf{94.94} \textpm 0.21 & \textbf{94.57} \textpm 0.23 & \textbf{67.83} \textpm 0.34 &\textbf{64.65} \textpm 0.53  & \textbf{69.42} \textpm 0.63 & \textbf{63.93} \textpm 0.59 & \textbf{88.04} \textpm 0.12 & \textbf{79.88} \textpm 0.24\\
\bottomrule
     \end{tabular}
     \vspace{-1mm}
\end{table*}

\begin{table*}[!t] 
    \centering
    \small
    \addtolength{\tabcolsep}{1.5pt}
    \caption{Node classification using other GNNs as local structure encoders.}
    \label{table.node-classification-gnns}%
    \vspace{-2mm}
    \begin{tabular}{@{}l|cc|cc|cc|cc@{}}
    \toprule
   \multirow{2}*{Methods} & \multicolumn{2}{c|}{DBLP} &
   \multicolumn{2}{c|}{IMDB} &  \multicolumn{2}{c|}{Freebase} & \multicolumn{2}{c}{AMiner}\\ 
      & Micro-F1 & Macro-F1 & Micro-F1 & Macro-F1& Micro-F1 & Macro-F1 & Micro-F1 & Macro-F1 \\\midrule\midrule
\method{Transformer} & 93.88 \textpm 0.66 &93.35 \textpm 0.58 & 67.21 \textpm 0.77 & 64.70 \textpm 0.65 & 67.89 \textpm 0.39 & 63.35 \textpm 0.46 & 85.72 \textpm 0.43 & 74.15 \textpm 0.28 \\
\midrule
\model-Adj & \uline{94.57} \textpm 0.25 & \uline{94.07} \textpm 0.30 & 67.98 \textpm 0.27 & 63.68 \textpm 0.58 
& \uline{68.98} \textpm 0.26 & \textbf{64.64} \textpm 0.49 & 86.82 \textpm 0.43 & 77.77 \textpm 0.77 \\
\midrule
\model-SAGE & 94.26\textpm0.09 & 93.85 \textpm 0.10 & \uline{68.02} \textpm 0.57 & \uline{64.15} \textpm 0.41 & 68.64 \textpm 0.44 & 64.12 \textpm 0.50 & 87.50 \textpm 0.35 & 79.06 \textpm 0.63
\\ 
\model-GIN  & 94.37 \textpm 0.24 & 94.01 \textpm 0.26 & 67.41 \textpm 0.41 & 63.55 \textpm 0.49 & 68.87 \textpm 0.83 & \uline{64.23} \textpm 0.67 & 87.72 \textpm 0.25 & 79.32 \textpm 0.50\\
\model-GCN    &\textbf{95.11} \textpm 0.27 & \textbf{94.75} \textpm 0.29 & \textbf{68.50} \textpm 0.37 & \textbf{65.59} \textpm 0.34 & \textbf{69.42} \textpm 0.63 & 63.93 \textpm 0.59 & \textbf{88.04} \textpm 0.12 & \textbf{79.88} \textpm 0.24
\\
 \bottomrule
     \end{tabular}
     \vspace{-2mm}
\end{table*}

\stitle{Baselines.} \label{baseline}
To comprehensively evaluate the proposed \model\ against the state-of-the-art approaches, we consider a series of baselines from three main categories, \emph{Homogeneous GNNs} (including \method{GCN} \cite{kipf2016semi}, \method{GAT} \cite{velivckovic2017graph}, \method{Transformer} \cite{vaswani2017attention}), \emph{Meta-path based HGNNs} (including \method{RGCN} \cite{schlichtkrull2018modeling}, \method{HetGNN} \cite{zhang2019heterogeneous}, \method{HAN} \cite{wang2019heterogeneous}, \method{GTN} \cite{yun2019graph}, \method{MAGNN} \cite{fu2020magnn}) and \emph{Meta-path free HGNNs} (including \method{RSHN} \cite{zhu2019relation}, \method{HetSANN} \cite{hong2020attention}, \method{HGT} \cite{hu2020heterogeneous}, \method{SimpleHGN} \cite{lv2021we}). 
We provide further details for these baselines in Appendix~\ref{app.baseline}.

\stitle{Settings and parameters.}
We conduct multi-class node classification on \emph{DBLP}, \emph{Freebase}, and \emph{AMiner}, while multi-label node classification on \emph{IMDB}.
For each dataset, we randomly split the nodes with the proportion of 24:6:70 for training, validation and test. 
In particular, for datasets \emph{DBLP} and \emph{IMDB}, we follow the standard split from HGB \cite{lv2021we}, and evaluate our model through the online leaderboard\footnote{https://www.biendata.xyz/hgb/}.
Micro-F1 and Macro-F1 are employed as metrics to evaluate the classification performance.
All experiments are repeated for five times, and we report the averaged results with standard deviations. We provide further details of hyper-parameter settings for both the baselines and \model\ in Appendix~\ref{app.hyperpara}.
% \zeminC{说明multi-class和label}

% For each dataset, we split node labels according to 24\% for training, 6\% for validation and 70\% for test. For \emph{DBLP} and \emph{IMDB}, we use the standard dataset released by HGB, and evaluate our model through online website of HGB\footnote{https://www.biendata.xyz/hgb/}. For \emph{AMiner} and \emph{Freebase}, we randomly split the node labels for evaluation. 
% Micro-F1 accuracy and Macro-F1 accuracy are employed as metrics to evaluate the performance. All experiments are repeated for five times, and we report the averaged results with standard deviations. We provide further details of parameters of baselines and \model\ in Appendix~\ref{app.hyperpara}.

\subsection{Performance Evaluation}

% We conduct node classification task on the four HIN benchmark datasets for comparison and we report the graph classification accuracy from two metrics, \ie, Micro-F1 and Macro-F1. 
We conduct node classification on the four benchmark datasets.
For our \model, we employ \method{GCN} \cite{kipf2016semi} as the local structure encoder. To evaluate the flexibility of local structure encoder, we further adopt \method{GraphSAGE}, \method{GIN}, and non-parameter \method{Adj} as shown in Eq.~\eqref{eq.local_adj}, as local structure encoders for comparison. 

\stitle{Node classification.}
% We compare our model with all the baselines on four HIN datasets. 
For \emph{DBLP} and \emph{IMDB}, as we adopt the standard settings for experiments, we directly borrow the results published in HGB leaderboard for comparison. 
For the other two datasets, we follow the default hyper-parameter settings for all the baselines in their literature and further tune them according to their validation performance. 
We report the performance comparison in Table~\ref{table.node-classification-all}, and observe that \model\ can outperform all the baselines.

In particular, we make the following observations.
Firstly, the standard \method{Transformer} can achieve competitive performances against HGNNs in most cases, which shows the effectiveness of context-based global attention mechanism for representation learning on HINs.
Secondly,  \model\ outperforms both standard \method{Transformer} and \method{GAT} by a large margin. 
% This demonstrates the feasibility of combining local structural information with a global attention mechanism on HINs. \zeminC{前面这个结论，Transformer不也是这样做的？} 
This demonstrates the effectiveness and necessity of our local structure encoder and heterogeneous relation encoder to capture both the underlying contexts and semantics for node representation learning.
Standard \method{Transformer} fails to capture contextual structures and the heterogeneity with a simplified model, thus still suffers from the overfitting issue, while \method{GAT} fails to learn heterogeneity similarity and to interact with long-range nodes on the graph.
% in the context directly without global attention. 
%\model\ can alleviate the shortage of both structural and heterogeneous information for Transformer architecture to form a more comprehensive representation;
Thirdly, it is interesting that no significant differences exist between the performance of meta-path based HGNNs and meta-path free HGNNs, showing that modeling heterogeneity in a learnable schema without prior knowledge might be a more practical manner. In light of this, our heterogeneous relation encoder might be a more beneficial exploration for heterogeneity on HINs to capture the underlying semantics between nodes.
% of modeling heterogeneity in a manner of position encoding.
% Thirdly, there are no significant advantages from meta-path based HGNNs compared with meta-path free HGNNs. Modelling heterogeneity in the learnable schema without prior knowledge may be more practical and our heterogeneous relation encoder is an exploration of modelling heterogeneity in a manner of position encoding.
Lastly, homogeneous GNNs are still powerful for representation learning on HINs, which demonstrates the effectiveness of neighborhood aggregation to capture the contextual structures on HINs.
% and the bottleneck of heterogeneous information utilization for HINs. 
In some sense, to effectively represent the nodes, the local structure information might be more important than high-order semantics which are usually denoted by some pre-defined semantic schemas, such as meta-paths. 

\stitle{Using other GNNs as local structure encoders.}
To demonstrate the flexibility of \model\ with different GNN backbones as the local structure encoder, we further utilize \method{GIN}, \method{GraphSAGE} and non-parameter \method{Adj} as the local structure encoders for comparison. 
For ease of comparison, we utilize a new split on datasets \emph{DBLP} and \emph{IMDB} for all the following experiments.
The performance results are reported in Table~\ref{table.node-classification-gnns}. We can observe that, \model\ is capable of outperforming Transformer with different structure encoders, demonstrating the flexibility of \model\ to work with different GNN backbones to capture the local structures, assisted with the exploitation of heterogeneity.
% to combine with different neighborhood aggregation architectures for capturing local structures, which is crucial to HIN representation learning. 
We also observe that, different GNN encoders contribute approximately to the overall performance, and even for \model-Adj which has no learnable parameters, it is interesting that it can still achieve promising performance, possibly due to the prominent power of neighborhood aggregation.
% Even without additional parameters, neighborhood aggregation can result in noticeable improvements for the Graph Transformer, and this \model-Adj may play a more important role in practice due to its efficiency.

\begin{table*}[!t] 
    \centering
    \small
    \caption{Impact of different global attention mechanisms on node classification.}
    \label{table.differ-attention}%
    \vspace{-2mm}
    \resizebox{1.0\textwidth}{!}{
    \begin{tabular}{@{}l|cc|cc|cc|cc@{}}
    \toprule
   \multirow{2}*{Methods} &
   \multicolumn{2}{c|}{DBLP} & \multicolumn{2}{c|}{IMDB} & \multicolumn{2}{c|}{Freebase} &  \multicolumn{2}{c}{AMiner} \\%\cline{2-9} 
& Micro-F1 & Macro-F1 & Micro-F1 & Macro-F1 & Micro-F1 & Macro-F1 & Micro-F1 & Macro-F1\\ 
\midrule
\method{GAT} &93.93 \textpm 0.19 & 93.47 \textpm 0.22 & 65.32 \textpm 1.13 & 59.69 \textpm 3.53 & \uline{69.04} \textpm 0.58 & 59.28 \textpm 2.56 & 84.92 \textpm 0.68 & 74.32 \textpm 0.95\\
\midrule
\method{\model-GAT} & 94.54 \textpm 0.20 & 94.05 \textpm 0.21 & 67.98 \textpm 0.23 & 64.51 \textpm 0.75 & 68.00 \textpm 0.66 & 63.29 \textpm 1.22 & \uline{87.68} \textpm 0.13 & 78.53 \textpm 0.48\\     
\method{\model-Transformer} & \uline{95.01} \textpm 0.13 & \uline{94.61} \textpm 0.14 & \textbf{68.66} \textpm 0.42 & \textbf{65.99} \textpm 0.49 & 68.28 \textpm 0.41 & \uline{63.49} \textpm 0.36 & 87.43 \textpm 0.16 & \uline{78.87} \textpm 0.46\\    
\method{\model-GATv2} &\textbf{95.11} \textpm 0.27 & \textbf{94.75} \textpm 0.29 & \uline{68.50} \textpm 0.37 & \uline{65.59} \textpm 0.34 & \textbf{69.42} \textpm 0.63 & \textbf{63.93} \textpm 0.59 & \textbf{88.04} \textpm 0.12 & \textbf{79.88} \textpm 0.24\\ \bottomrule
    \end{tabular}}
    \vspace{-2mm}
\end{table*}

\subsection{Model Analysis} \label{sec.analysis}

We further analyze \model\ from several aspects for node classification, with \method{GCN}-backboned local structure encoder.

\stitle{Ablation study.}
To evaluate the contribution of each component in \model, we conduct an ablation study by comparing with several degenerate variants: 
(1) \emph{Transformer}: we use the vanilla Transformer architecture with the $D$-hop context sampling strategy;
(2) \emph{no LSE}: we remove the local structure encoder, and use the original features of nodes transformed by heterogeneous feature projection as the input node representations. 
(3)\emph{no HRE}: we remove the heterogeneous relation encoder and only calculate semantic similarities with their features, and this variant can be seen as a homogeneous node-level Graph Transformer; 
(4) \emph{no HRE \& LSE}: we remove both the local structure and heterogeneous relation encoders together.

\begin{figure}[t]
\begin{minipage}[t]{0.245\textwidth}
\centering
 \centerline{\includegraphics[scale=0.145]{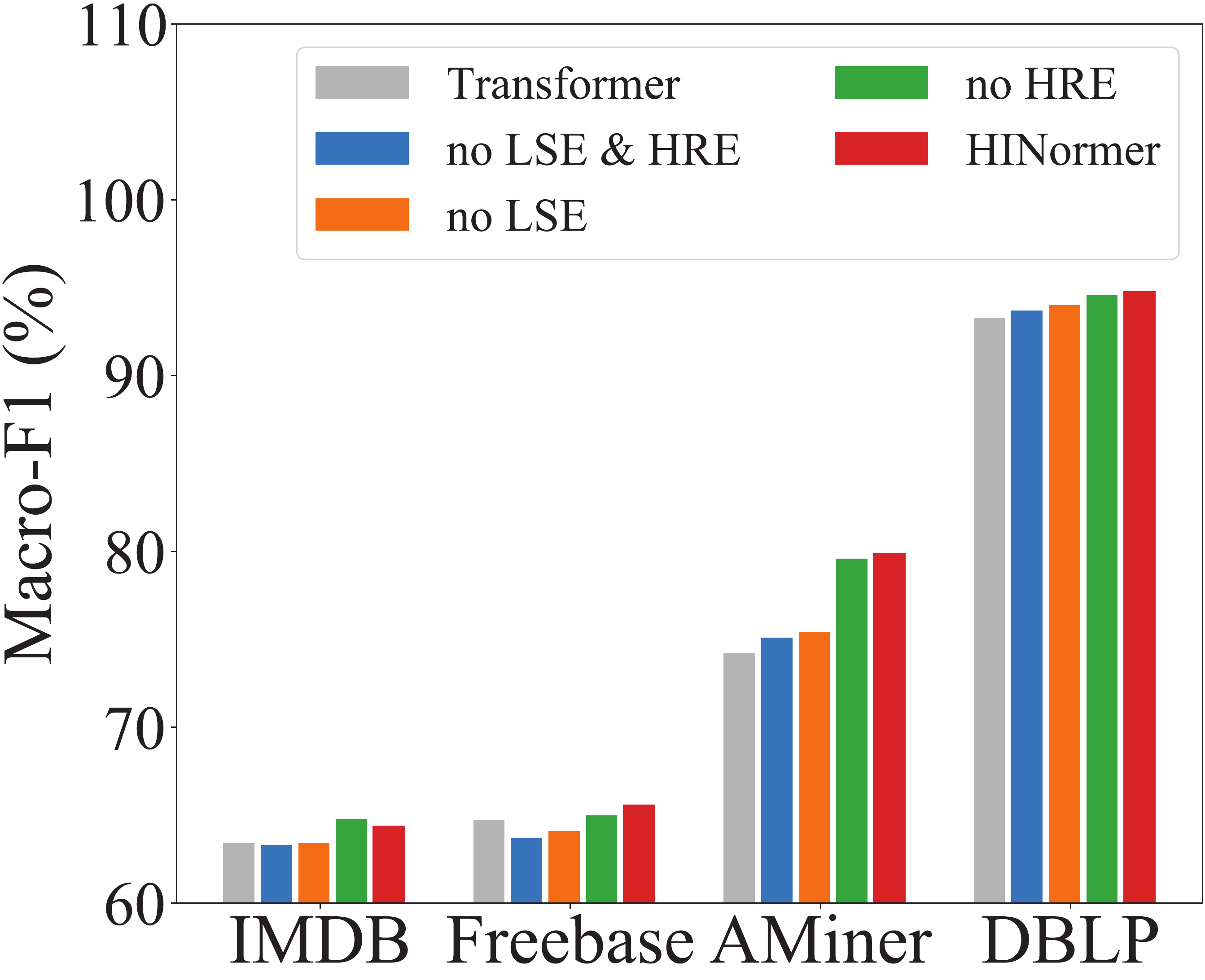}}
\vspace{-2mm}
\caption{Ablation study.}
\label{fig.ablation-study}
\end{minipage}% 
\begin{minipage}[t]{0.245\textwidth}
\centering
 \centerline{\includegraphics[scale=0.145]{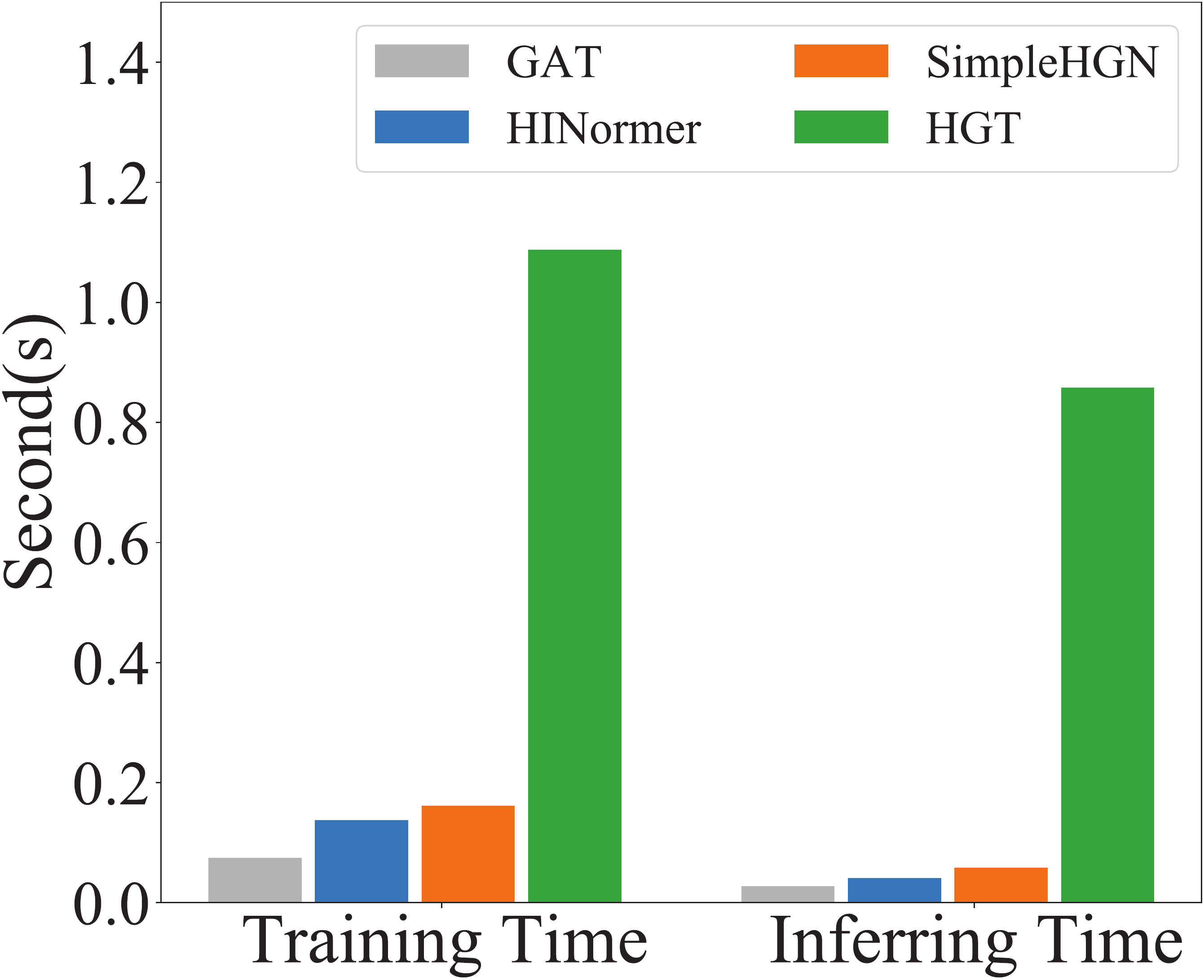}}
\vspace{-2mm}
\caption{Scalability study.}
\label{fig.scalability-study}
\end{minipage}
\vspace{-3mm}
\end{figure}

We show the results of ablation study in Fig.~\ref{fig.ablation-study}, and make the following observations. 
Firstly, without the local structure encoder, \model\ fails to encode the local contexts for node representations, resulting in a sharp performance degradation. It is necessary for the application of Graph Transformers on HINs to complement local structural information with appropriate modules, and neighborhood aggregation might be a promising choice.
Secondly, without heterogeneous relation encoder, the performance also generally decreases, demonstrating that the exploitation of semantic relations is beneficial to \model.
From another perspective, the improvement is not significant, thus how to effectively exploit the underlying semantic relations between nodes is still a burning question and we will continue to investigate in our future work.
% and that explicit heterogeneous information does not play a significant role in heterogeneous Graph Transformers. We have tried other methods to encode heterogeneous information explicitly, but none of them bring significant improvements. The more powerful heterogeneity modeling for the HIN Transformer will be left for future work.
Thirdly, without both HRE and LSE, the performance significantly decreases. Due to the utilization of \method{GATv2} \cite{brody2021attentive} with the simplified attention mechanism instead of the vanilla self-attention in \method{Transformer} to cope with the overfitting issue, \emph{no HRE \& LSE} can still outperform \emph{Transformer}.
% \model\ cannot utilize structure and heterogeneity of HIN so that the performance decreases obviously, but it still outperforms the standard Transformer due to overfitting caused by more parameters. 
Finally, the whole model \model\ can achieve the best performance, demonstrating its effectiveness for node representation learning on HINs.

\begin{figure}[t]
  \centering
  \vspace{-1mm}
  \subfigure[Number of Transformer layers]{
\includegraphics[width=0.46\linewidth]{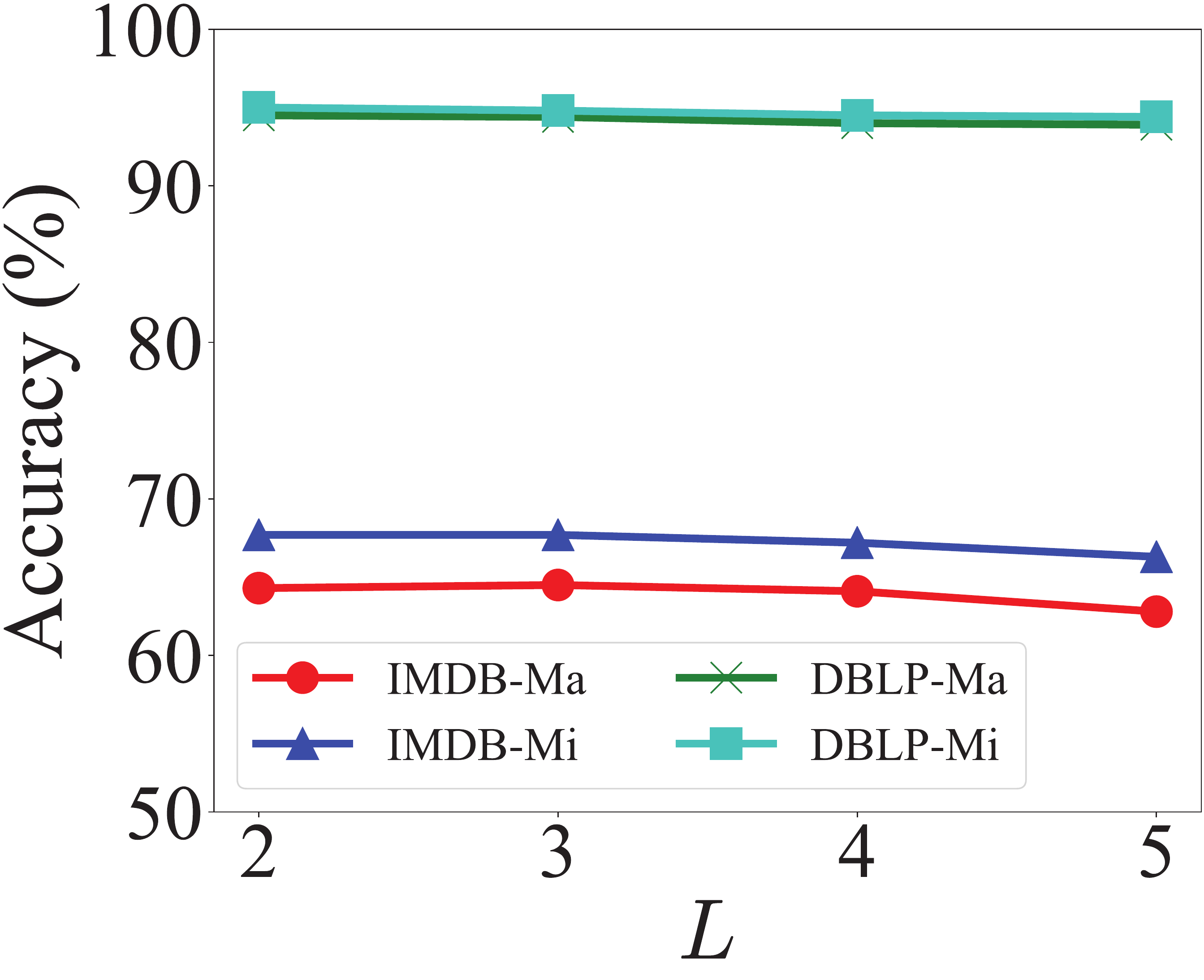}
   \vspace{-2mm}
   \label{fig.params-layer}
   }
  \subfigure[Sequence length S]{
  \includegraphics[width=0.46\linewidth]{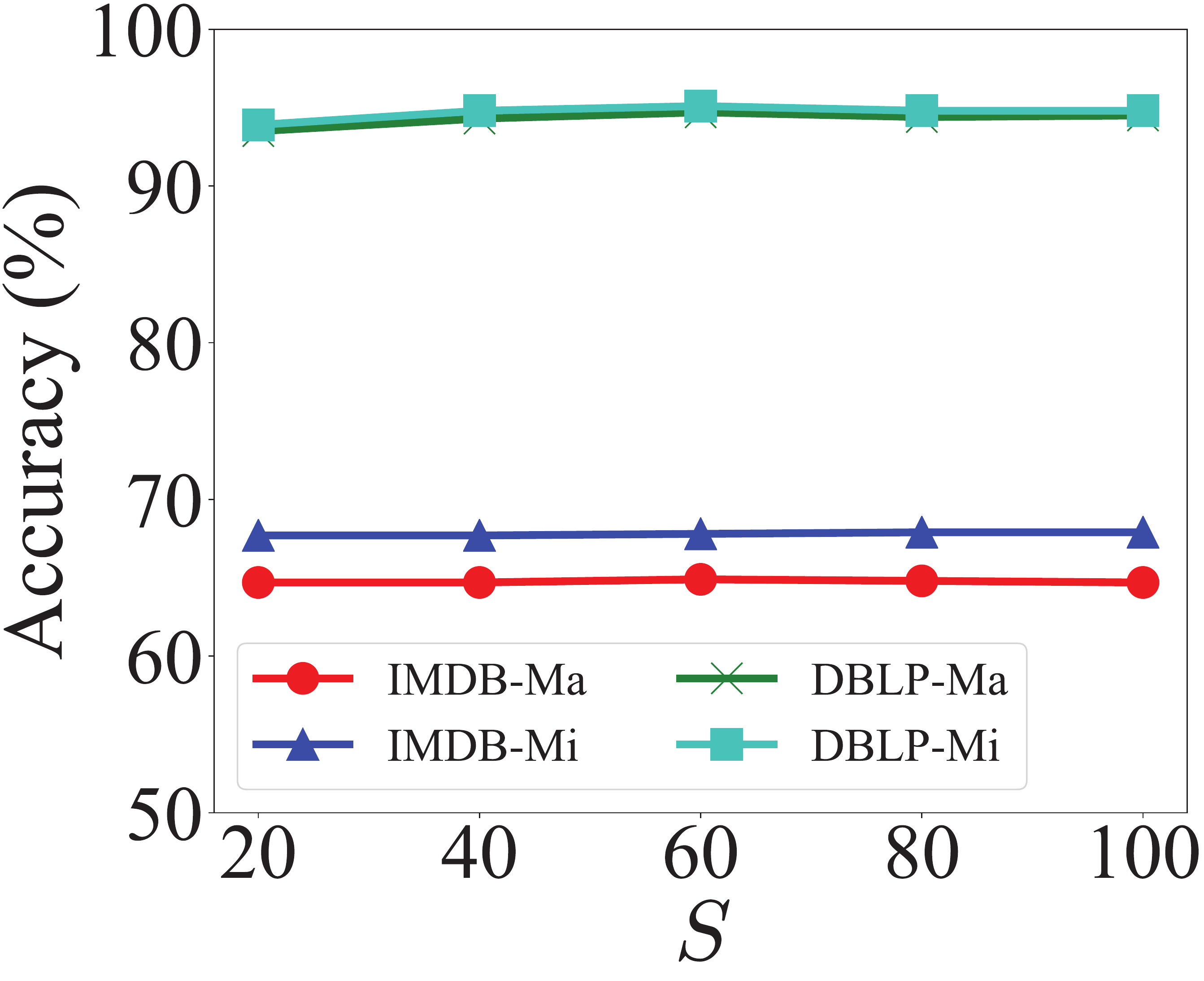}
  \vspace{-2mm}
  \label{fig.params-s}
  }
  \subfigure[hidden dimension $d$]{
  \includegraphics[width=0.46\linewidth]{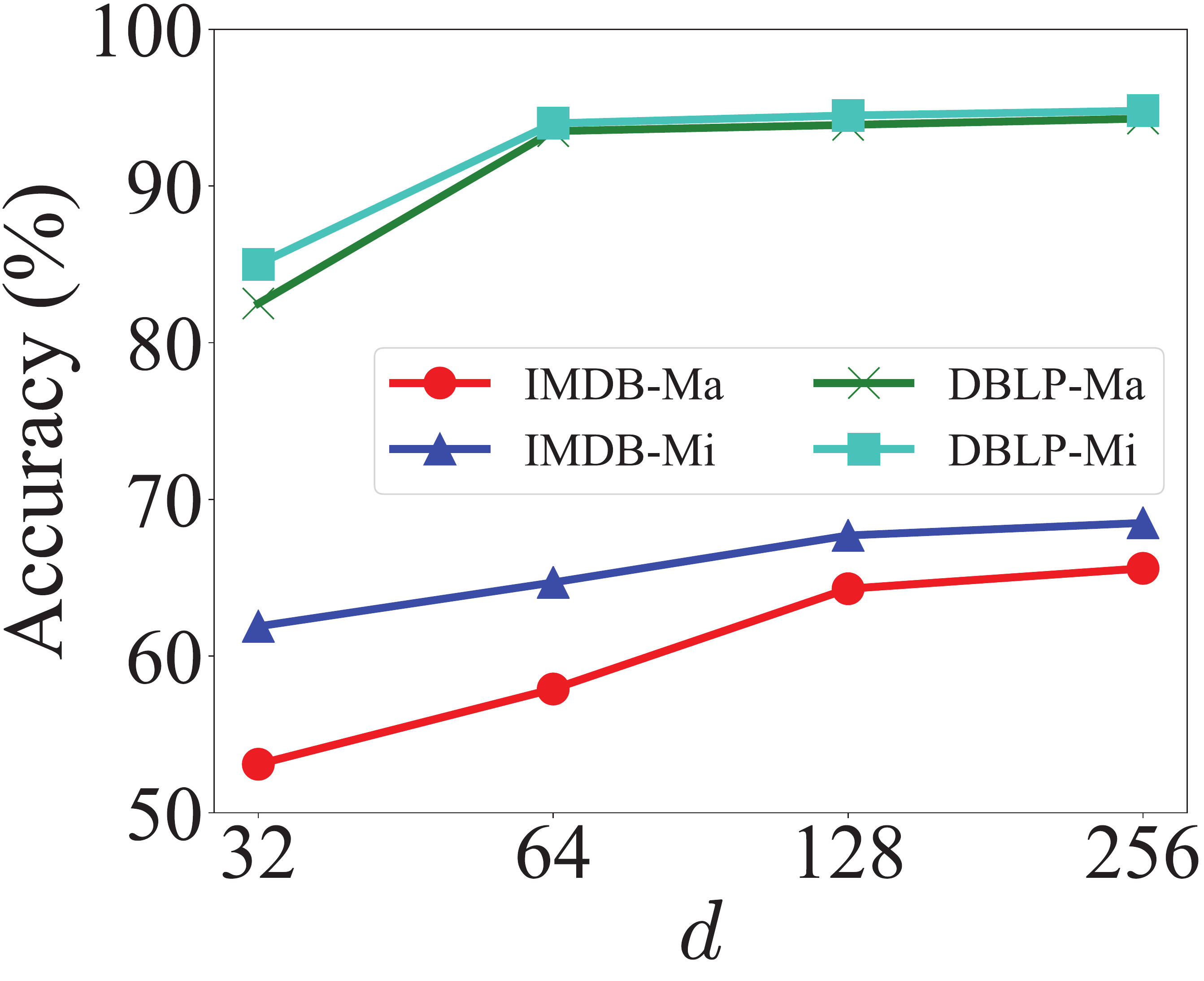}
  \vspace{-2mm}
  \label{fig.params-d}
  }
  \subfigure[Weight of heterogeneity similarity $\beta$]{
  \includegraphics[width=0.46\linewidth]{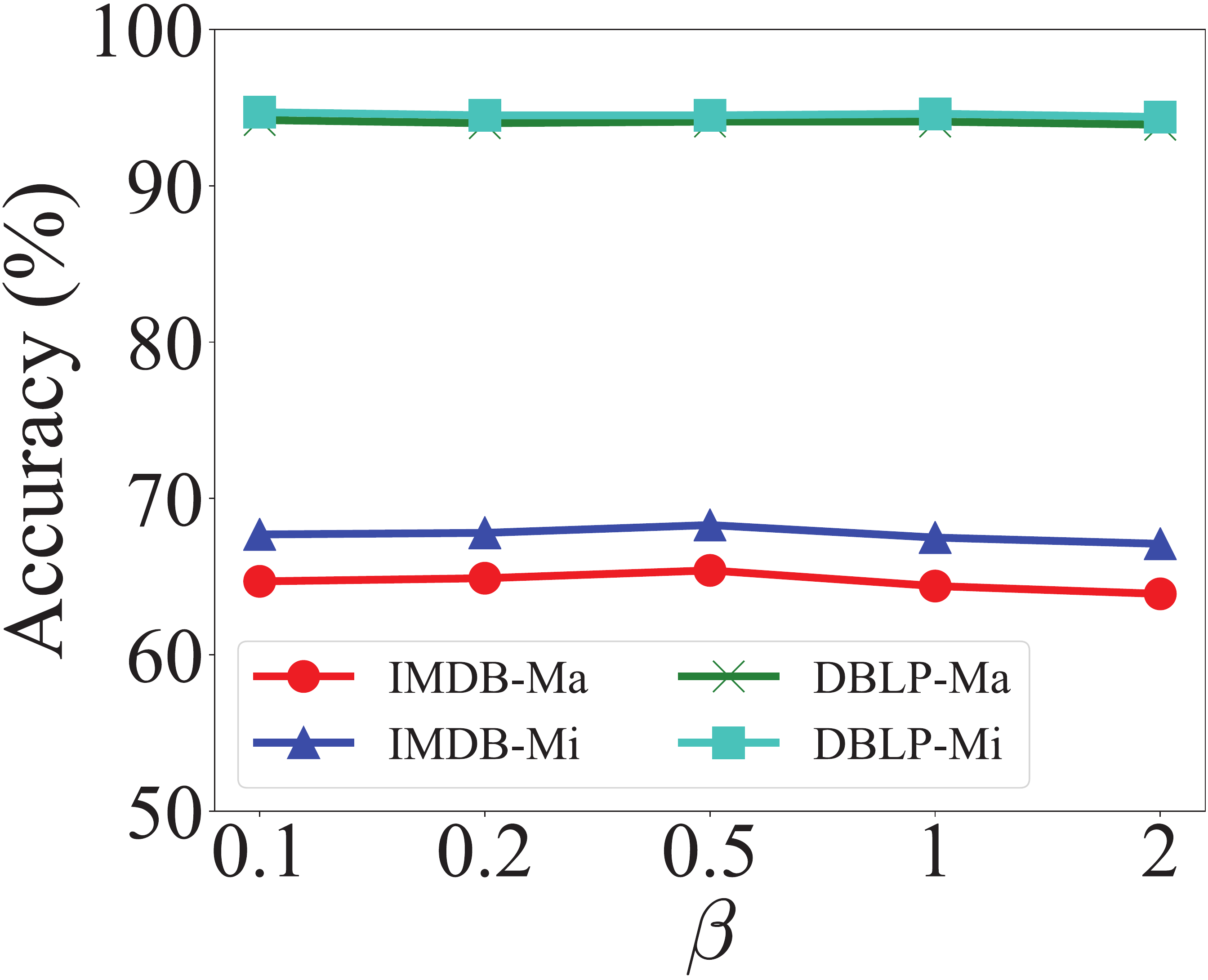}
  \vspace{-2mm}
  \label{fig.params-beta}
  }
\vspace{-3mm}
  \caption{Parameters sensitivity.} 
\label{fig.parameters}
\vspace{-3mm}
\end{figure}

\stitle{Impact of different attention mechanisms.}
% We replace the dot-product attention of Transformer with GATv2\cite{brody2021attentive} for more effective global attention mechanism. 
To explore the impact of different global attention mechanisms, we conduct experiments on \model\ with three different attention mechanisms including \method{GAT}, \method{Transformer}, and \method{GATv2}, and illustrate the performance comparison in Table~\ref{table.differ-attention}. Note that, the default attention mechanism in \model\ is \method{GATv2}.
We first observe that all three variants can generally outperform \method{GAT}, showing that it is the global attention mechanisms rather than the neighborhood-level attention that can help to achieve superior performance.
% We first observe that \model\ with all three global attention mechanisms outperform GAT with message-passing based attention mechanism. Employing global attention mechanism on HIN Transformer generates more discriminative node representations from direct message-passing in context range;
Secondly, though the three variants can achieve approaching performance, \model-\method{GATv2} performs more stably and can slightly beat the others across the four datasets with fewer parameters.
% the performances of \model-GATv2 are stable across different datasets and slightly better than other methods in most cases with less parameters than dot-product attention.

\stitle{Time comparison.}
We conduct node classification on \emph{DBLP} to evaluate the time cost of both training (per epoch) and inferring for several representative models for comparison, including \method{GAT}, \method{HGT}, \method{SimpleHGN} and \model.
% We test the both training and inferring time consumption of \method{GAT}, \method{HGT}, \method{SimpleHGN} and \model\
% for node classification on the \emph{DBLP} dataset. 
The results are shown in Fig.\ref{fig.scalability-study}. 
We observe that our \model\ is as efficient as \method{SimpleHGN} and much faster than \method{HGT}. Relative low time cost with fewer parameters enhances the availability of \model\ in different settings.
% endures the scalability of \model\ in practice.
% We measure the average time consumption of one epoch for each model, and our \model\ is as efficient as \method{SimpleHGN} with context global attention and much faster than \method{HGT}. Relative low time consumption with less parameters endures the scalability of \model\ in practice.

\stitle{Parameters Sensitivity.}
We evaluate the sensitivity of several important hyperparameters in \model, and show their impact in Fig.~\ref{fig.parameters}.
For the number of Transformer layers, smaller $L$ is generally better for the overall performance.
For the sequence length $S$ which denotes the context size, its change does not significantly impact the performance, and a small $S$ might be better due to the model efficiency.
% For the number of Transformer layers $L$ and sequence length $S$, preferring a relatively small value ensures the scalability of \model\ in application.
% For the sequence length (context size) $S$, 
%too small values may not provide sufficient context information while 
% too large values contribute large memory cost of calculation and less related nodes will bring noisy information that impairs the performance. 
%too small values may limit the model expressiveness, while too large values may result in overfitting. In particular, relative small values such as {2,3,4,5}.
For the hidden dimension of Transformer layer, better performance usually profits from a larger $d$, due to its stronger capacity to capture the various semantics stemming from the graph heterogeneity.
% , heterogeneous Graph Transformer needs a sufficient embedding size to capture the comprehensive relations on nodes in HIN,.
% For the hidden dimension of Transformer layer $d$, heterogeneous Graph Transformer needs a sufficient embedding size to capture the comprehensive relations on nodes in HIN,.
%too small values will fail the global attention mechanism.
For heterogeneous similarity weight $\beta$, moderate values such as [0.1,0.5] would benefit the performance.
%too small values may limit the effect of heterogeneous similarity in attention mechanism, while too large values may disturb the main feature-based similarity. Moderate values such as [0.1,0.5] would benefit the performance.

%% file: sec-conclusions.tex
\section{Conclusion}

In this paper, we investigate the problem of employing Graph Transformer on heterogeneous information networks for node representation learning. To achieve this, we propose a novel model named \model, which capitalizes on a self-attention mechanism assisted by two key components, \ie, a local structure encoder and a heterogeneous relation encoder, to capture both the structural and heterogeneous information for node representation learning. Extensive experiments on four benchmark datasets demonstrate the effectiveness of our proposed \model.
% \zemin{assisted by transformer}

% In this paper, we investigate the possibility of applying global attention Transformer architecture to HINs and propose a complete effective paradigm for the design of heterogeneous Graph Transformer with simplified architecture and efficient context sampling strategy. To capture both structural and heterogeneous information for heterogeneous Graph Transformer, two novel encoder module \ie, local structure encoder and heterogeneous relation encoder, are employed in \model. Extensive experiments on four benchmark datasets demonstrate the effectiveness of our proposed \model.

%% file: sec-appendix.tex
\section*{Appendices}
\renewcommand\thesubsection{\Alph{subsection}}
\renewcommand\thesubsubsection{\thesubsection.\arabic{subsection}}

\subsection{Further Details of Datasets} \label{app.datasets}
\stitle{Datasets.} 
We employ a total of four real HIN benchmark datasets, including two academic citation datasets and two movie-related datasets. 
The academic citation datasets include \emph{DBLP}and \emph{AMiner}, while the movie-related datasets include \emph{IMDB} and  \emph{Freebase}.
\begin{itemize}
    \item \emph{DBLP} is a bibliographic dataset in computer science, the dataset includes papers published between 1994 to 2014 of 20 conferences in 4 research fields. There are four types of nodes including authors(A), papers(P), terms(T) and venues (V). And we directly use the split dataset from HGB \cite{lv2021we}.
    \item \emph{AMiner} is also an academic network, we uses a subgraph of the original dataset including papers of four classes. There are three types pf nodes including papers(P), authors(A) and references(R).
    \item \emph{IMDB} is a website about movies and related information, the only multi-label dataset includes movies from Action, Comedy, Drama, Romance and Thriller classes.  There are four types of nodes including movies(M), directors(D), actors(A) and keywords(K). And we directly use the split dataset from HGB.
    \item \emph{Freebase}\cite{bollacker2008freebase} is a huge knowledge graph, and we use a subgraph of 4 genres of entities including movies(M), actors(A), directors(D) and writers(W).
\end{itemize}

\subsection{Further Details of Baselines} \label{app.baseline}
To comprehensively evaluate the proposed \model against the state-of-the-art approaches, we consider a series of baselines from three main categories, \emph{Homogeneous GNNs}, \emph{Meta-path based HGNNs} and \emph{Meta-path free HGNNs}.

\noindent (1) \emph{Homogeneous GNNs}
\begin{itemize}
  \item \method{GCN}: \method{GCN} \cite{kipf2016semi} depends on the key operation of neighborhood aggregation to aggregate messages from the neighboring nodes to form the node representations. In particular, it employs a mean-pooling to aggregate the neighborhood information.
  \item \method{GAT}: \method{GAT} \cite{velivckovic2017graph} replaces the average aggregation from neighbors with weighted aggregation from addition attention mechanism. And it also employs multi-head attention technique to improve the performance.
  \item \method{Transformer}: \method{Transformer} \cite{vaswani2017attention} employs the same context sampling strategy to conduct node-level representation learning with standard Transformer architecture.
\end{itemize}

\noindent (2) \emph{Meta-path based HGNNs}
\begin{itemize}
  \item \method{RGCN}: \method{RGCN} \cite{schlichtkrull2018modeling} extends \method{GCN} to multiple edge types graphs, decomposing the heterogeneous graph convolution into two stages. For each node, the first step performs mean aggregation on specific edge type graph, and the second step aggregates the representations from all edge types.
 \item \method{HetGNN}: \method{HetGNN} \cite{zhang2019heterogeneous} firstly aggregates content features of nodes inside each neighborhood generated by type-specific random walks with a Bi-LSTM aggregator to get the type-specific embedding, then employs the attention mechanism on all type-specific to generate the final embedding vector for each node.
 \item \method{HAN}: \method{HAN} \cite{wang2019heterogeneous} also employs a hierarchical attention mechanism which includes the node-level attention for meta-path based neighbor aggregation and the semantic-level attention for meta-path related semantic aggregation. Both node-level and semantic-level attention are implemented by \method{GAT}.
 \item \method{GTN}: \method{GTN} \cite{yun2019graph} replaces manual selection on meta-paths with automatic learning process of meta-paths, achieving the goal through sub-graph selection and matrix multiplication with learnable \method{GCN}s.
 \item \method{MAGNN}: \method{MAGNN} \cite{fu2020magnn} is based on \method{HAN} and further takes use of all nodes in a meta-path instance rather than only the nodes of the two endpoints.
\end{itemize}

\noindent (3) \emph{Meta-path free HGNNs}
\begin{itemize}
  \item \method{RSHN}: \method{RSHN} \cite{zhu2019relation} firstly builds coarsened line graph to obtain embeddings of different edge types, and then uses a Message-Passing neural network to propagate both node and edge features.
\item \method{HetSANN}: \method{HetSANN} \cite{hong2020attention} employs type-specific \method{GAT} layers for the aggregation of local information to capture heterogeneous information.
  \item \method{HGT}: \method{HGT} \cite{hu2020heterogeneous} proposes a heterogeneous Transformer-like attention mechanism for neighborhood aggregation which is not global attention, using type-specific parameters to characterize the heterogeneous attention over each edge. It also proposes a type-aware sampling method to tackle large-scale HIN.
  \item \method{SimpleHGN}: \method{SimpleHGN} \cite{lv2021we}  proposes a simple but strong baseline \method{GAT}-based model which considers both the edge type embedding and node embeddings to calculate the attention score. It constructs the Heterogeneous Graph Benchmark (HGB) to standardize the process of HIN representation learning, making a solid contribution on the healthy development of HINs.
\end{itemize}

\subsection{Hyperparameters Settings} \label{app.hyperpara}
\stitle{Baselines.} As we adopts results of baseline published in HGB for \emph{DBLP} and \emph{IMDB}, we only report the hyperparameter settings of \emph{Freebase} and \emph{AMiner}. For meta-path based HGNNs, we use "M-A-M", "M-D-M" and "M-W-M" as meta-paths for \emph{Freebase}, "P-A-P" and "P-R-P" as meta-paths for \emph{AMiner}. \par
For baseline \method{GCN} \cite{kipf2016semi}, we set $d$ = 64 for both datasets. We set
$L$ = 3 for \emph{Freebase}, and $L$ = 4 for \emph{AMiner}.
For \method{GAT} \cite{velivckovic2017graph}, we set $d$ = 64, $n_h$ = 8, $L$ = 3, $s$ = 0.05 for both datasets.
For \method{RGCN} \cite{schlichtkrull2018modeling}, we set $d$ = 32, $L$ = 5 for both datasets.
For \method{HetGNN} \cite{zhang2019heterogeneous},  We set $d$ = 128, and batch size as 200 for all datasets. For random walk, we set walk length as 30
and the window size as 5.
For \method{HAN} \cite{wang2019heterogeneous}, we set $d$ = 8, $d_a$=128, $n_h$ = 8, $L$ = 3 for both datasets.
For \method{MAGNN} \cite{fu2020magnn} we set $d$ = 64, $d_a$=128, $n_h$ = 8, batch size as 8 and number of neighbor samples as 100 for both datasets.
For \method{RSHN} \cite{zhu2019relation}, we set $d$ = 32, $L$ = 3 for both datasets.
For \method{HGT} \cite{hu2020heterogeneous}, we use layer normalization in each layer, and set $d$ = 64, $n_h$ = 8, $L$ = 3 for both datasets. 
For \method{SimpleHGN} \cite{lv2021we}, we set $d$ = $d_e$ = 64, $n_h$ = 8, $\beta$ = 0.05, $L$ = 3 for both datasets. We set $s$ = 0.05 for \emph{AMiner} and $s$ = 0.1 for \emph{Freebase}.

\stitle{Our model.}
We use Adam\cite{kingma2014adam} as optimizer, LeakyReLU with negative slope $s$ = 0.2 as activation function and adopt the ReduceLROnPlateau scheduler of Pytorch on validation loss, and early-stopping with patience of 50 is adopted to prevent overfitting. The learning rate is set to 1e-4, dropout rate is set to 0 for \emph{freebase} and 0.5 for other datasets. The range of sequence length $S$ is [10,200], the hidden dimension $d$ is set to 256 and the number of head $n_h$ is set to 2 for all datasets. The number of layers for local structure encoder $K_s$ and heterogeneous relation encoder $K_h$ is set to 5 for \emph{DBLP}, 4 for \emph{IMDB} and 3 for others. The search space of number of Transformer layers is \{2,3,4,5\} and \{0.1, 0.2, 0.5, 1, 2\} for heterogeneous weight $\beta$.